\documentclass[12pt, draftclsnofoot, onecolumn]{IEEEtran}
\IEEEoverridecommandlockouts
\usepackage{cite}
\usepackage{amsmath,amssymb,amsfonts,bm}
\usepackage{algorithmic}
\usepackage{algorithm,float}
\usepackage{graphicx,subfigure}
\usepackage{textcomp}
\usepackage{xcolor}
\usepackage{multirow}
\usepackage{balance}
\usepackage{colortbl}
\usepackage{url}
\def\BibTeX{{\rm B\kern-.05em{\sc i\kern-.025em b}\kern-.08em
    T\kern-.1667em\lower.7ex\hbox{E}\kern-.125emX}}

\begin{document}

	\title{Multi-Objective DNN-based Precoder for \\MIMO Communications 
}

\author{%
	\IEEEauthorblockN{Xinliang Zhang, \textit{Student Member}, 
	and Mojtaba Vaezi, \textit{Senior Member}, \textit{IEEE}\\}
	
}
\maketitle
\begin{abstract}
This paper introduces a unified deep neural network (DNN)-based precoder for two-user multiple-input 
multiple-output (MIMO) networks with five objectives: 
data transmission, energy harvesting, simultaneous wireless information 
and power transfer, physical layer (PHY) security, and multicasting. 
First, a rotation-based precoding  is developed to solve the above problems independently. Rotation-based precoding is new precoding and power allocation that 
beats existing solutions in PHY security
and multicasting and is reliable in  different antenna settings.     Next, a DNN-based precoder is designed to unify 
the solution for all objectives. The proposed DNN concurrently learns the solutions given by conventional methods, i.e., 
	analytical  or rotation-based solutions. A binary vector is designed as an input 
feature to distinguish the objectives. Numerical results demonstrate that, compared to the conventional solutions, 
the proposed DNN-based 
precoder reduces on-the-fly   
computational complexity  more than an order of magnitude  while reaching near-optimal performance  ($99.45\%$ of the averaged optimal solutions). 
The new precoder is also more robust to the variations of the numbers of antennas at the receivers.  
\end{abstract}
\section{Introduction}\label{sec_intro}

{\let\thefootnote\relax\footnotetext{
%
%
%
		
	The authors are with the Department of Electrical and Computer Engineering, Villanova University, Villanova, PA 19085, USA
	(Email:  \{xzhang4, mvaezi\}@villanova.edu\})
		
}}

 Wireless communication faces unprecedented  challenges  in terms of diverse 
 {\color{black} objectives {(e.g., throughput, energy efficiency, security, 
	and delay}) and emerging
applications} (e.g., Internet of things 
 {(IoT)}, wearables, drones, etc.). 
As a recent example, with a daily average data rate over 16.6 Gigabytes, the 
communication traffic for in-home data usage during the coronavirus 
(COVID-19) outbreak 
in March 2020 has increased 18 percent compared to the same period in 
2019 
\cite{statista2020internet}. 
Such multifaceted challenges are conventionally addressed  
separately  in the physical layer (PHY) because it is 
not possible to come up with one optimal solution satisfying all of those diverse  
{\color{black} and, at 
times,} clashing requirements and 
objectives. 
However, in practice, many 
of those objectives should 
be satisfied simultaneously in some  applications, e.g., in IoT 
devices which have limited computational resources but need to harvest energy for their transmission).
Conventional solutions may even differ only if the number of antennas at the users. 

\begin{figure}[t]
	\centering
	\includegraphics[width=0.68\textwidth]{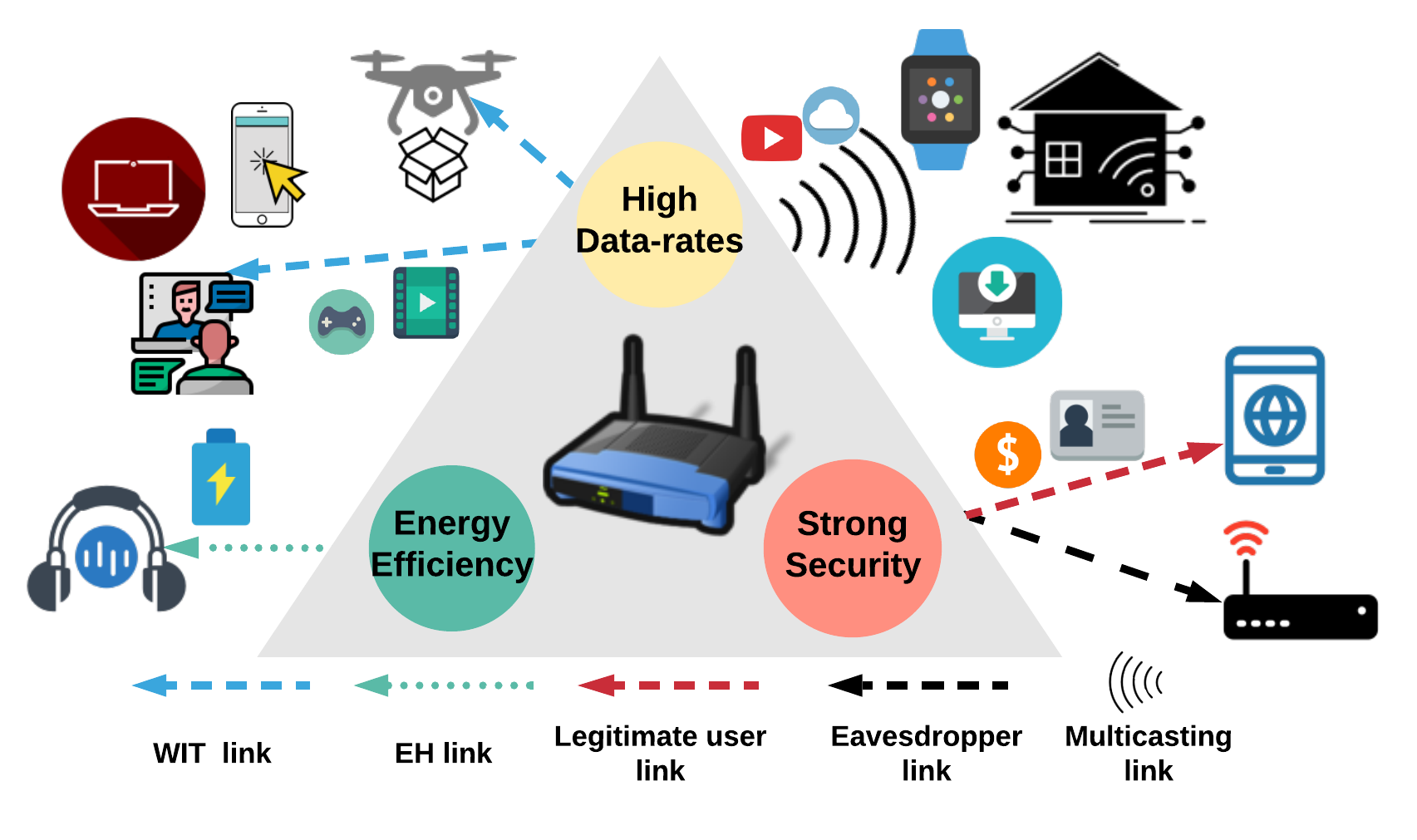}
	\caption{A system with multiple communication 
		services.}
	\label{fig:concept} 
\end{figure}

Motivated by the above, 
a streamlined
system (illustrated in Fig.~\ref{fig:concept}) is unified with prolific 
transmission functions: high data rates, strong security, and 
efficient energy exploitation.
The integrated transmission system is required for three facets (categories of tasks) 
simultaneously: 1) data transmission 
such 
as wireless information transmission (WIT) and multicasting; 2) green 
communication, including energy harvesting (EH) and  
simultaneous 
	wireless information and power transfer
(SWIPT); 3) secure communication, e.g., physical layer (PHY) security.
This full-featured system motivates us to consider the following  
question: How can we 
integrate integrate all of these facets  into one system with an acceptable or 
even better performance?


To answer this question, it is enlightening to understand the current approaches to address those objectives. As an essential part of the multiple-input and multiple-output  (MIMO) communication systems, 
\textit{precoding} and \textit{power allocation} schemes (or equivalently, transmit covariance matrix design) 
are typically used to address each of those facets independently of the others.
More specifically, for each of the five objectives we mentioned (i.e., WIT, EH, SWIPT, PHY security, 
and  multicasting), one or more independent solutions are developed in the literature (see in Table~\ref{tab_modes}). 
For some objective, such as WIT, an 
optimal closed-form solution is known, which is obtained via celebrated 
singular value decomposition (SVD) and water-filling 
\cite{cover2012elements}. Linear precoding and power allocation solutions 
for EH and SWIPT can be 
found in \cite{zhang2013mimo} and \cite{ rostampoor2017energy}.
For others, such as PHY security in the MIMO wiretap channel, only sub-optimal or iterative solutions are known in general 
\cite{fakoorian2012optimal,li2013transmit,zhang2019rotation} as the problem is not convex. 
Among them, generalized singular value decomposition 
(GSVD)-based precoding \cite{fakoorian2012optimal} is fast, sub-optimal solution whereas  alternating 
optimization with water-filling (AO-WF) 
\cite{li2013transmit}  has better performance but 
 requires much more time.  Yet, those methods may not 
be 
close to the 
capacity in some antenna settings \cite{vaezi2017journal, 
	zhang2019rotation}.  
Lastly, multicasting is a min-max fair problem to enlarge the 
transmission rate for all users. 
In  the multiple-input single-output (MISO) case,  
semidefinite relaxation 
(SDR) techniques yield a closed-form solution 
\cite{sidiropoulos2006transmit}. In the MIMO case, a cyclic alternating 
ascent  (CAA) linear precoding is 
proposed in \cite{zhu2012precoder}.

\begin{table}[t]
	\caption{The Desired Objectives and Existing Solutions}
	\label{tab_modes}
	\centering
	\begin{tabular}{c c c c}
		\hline
		Configuration & Objective Function  & 
		Reference \\ \hline
		\rowcolor[HTML]{EFEFEF} 
		 $\mathcal{O}_1$   &    WIT   & \cite{cover2012elements} \\  
		 $\mathcal{O}_2$    &    EH   & \cite{zhang2013mimo}\\
		\rowcolor[HTML]{EFEFEF} 
		 $\mathcal{O}_3$ &  SWIPT &  \cite{zhang2013mimo, 
			rostampoor2017energy}  \\
		 $\mathcal{O}_4$    &  PHY Security  & 
		\cite{fakoorian2012optimal, li2013transmit,
			vaezi2017journal,zhang2019rotation} \\ 
		\rowcolor[HTML]{EFEFEF} 
		 $\mathcal{O}_5$     &    Multicasting & 
		\cite{sidiropoulos2006transmit, 
			zhu2012precoder}\\		     
		\hline
	\end{tabular}
\end{table}
  
 It is seen that various different approaches are used to design precoder for the problems listed in Table~\ref{tab_modes} and, yet, some of them are not effective in all antenna settings. 
 Among those problems, $\mathcal{O}_3$ to $\mathcal{O}_5$ are more challenging.  In the first part of this paper, we 
  apply \textit{rotation-based precoding} (RP) to the latter three problems.
  This approach uses one method of solution for all those three problems.\footnote{RP can be applied to all of the five problems listed in Table~\ref{tab_modes}. However,  RP has no advantage over the existing solutions for WIT and EH as analytical solutions are available for them.} More importantly, in general, it results in a better performance for these 
  objectives when compared with existing methods.  RP can be applied to all of those problems to unify the optimization approach. However, the optimization problems corresponding to those objectives are still solved individually.   

In the second part of this paper, we introduce a unified deep neural network (DNN)-based precoder to solve optimization problem corresponding to all of those five objectives ($\mathcal{O}_1$ to $\mathcal{O}_5$) at once. 
This will settle the question we raised earlier in this paper. 
The new question is how we can ``teach'' a DNN 
\cite{lecun2015deep} to concurrently and effectively ``learn'' all objectives together? 
To this end, we utilize a 
supervised DNN to learn from the solutions given by the RP by using the backpropagation 
algorithm which updates internal parameters from presentation layers 
\cite{lecun2015deep}. We introduce an input feature to distinguish the objectives, and we interpret ``learn'' as the act of choosing the best precoder.
DNN is a good ``learner'' due to its sensitivity to the same types of input 
and output pairs, even if mathematical models/solutions for those pairs
are totally different. DNN-based precoding can realize 
unification by regulating all the input objectives with a same structure.

Before this work, DNN has separately been applied to many communications problems independently. To name a few, in \cite{shanin2019rate}, DNN is 
employed to model a Markov chain 
to obtain the rate-energy region of SWIPT in practical EH circuits. In \cite{fritschek2019deep},  an 
autoencoder is proposed in which DNN learns the optimal mapping from encoder to 
decoder for PHY 
security. In \cite{zhang2019deep}, a DNN-based 
precoder for wiretap channel is considered for specific antenna 
settings. 
We do not expect that supervised DNN to significantly 
surpass conventional mathematical methods in communications; 
nevertheless, DNN holds promise for 
many front-end technologies in complex scenarios \cite{luo2019machine}, 
such as spectrum intelligence which  deceptively manages the radio 
resource \cite{lee2018resource, doan2019power, d2018learning};  
transmission intelligence which focuses on reaching  channel estimation and 
characterization \cite{o2017introduction, li2019deep};  network intelligence 
which 
enhances the communication quality in a system-level
\cite{fan2014self}.



\subsection{Motivation and Contribution}\label{sec_intro_sub_ctb}
Apart from unifying precoder design for  multi-objective systems discussed earlier,  computation efficiency is another reason that motivates us to investigate a  DNN-based precoder. 

DNN is a universal function
approximator \cite{hornik1989multilayer}  which has 
achieved a remarkable capacity of algorithm learning \cite{zoph2016neural}. 
DNN is essentially a set of filters that is applied repeatedly 	 to batches 
of the input. The network owns fixed times of convolution operation on 
weight matrices which can avoid the endless loop of iteration. In other words, 
DNN is able to achieve high resource utilization (e.g., matrix rather than 
vector 
operations on GPU),  which in turn can substantially reduce the 
computation 
costs without sacrificing accuracy. 
As a data-driven technique, DNN is usually arranged offline and only needs to 
be performed  once.
In this paper, we design a unified precoder based on a DNN architecture 
which explores a different way of thinking for wireless communication systems.
Our main contributions are: 
\begin{itemize}
	\item We introduce rotation-based precoding and power allocation  for multiple objectives,
	including WIT, EH, SWIPT, PHY security, and multicasting.
	To the best of 
	our knowledge, there is no such work linking so many functions into one model. The rotation-based model can 
	parameterize any covariance matrices, which makes it possible to train a 
	DNN in a unified manner.
	\item 
Rotation-based precoders  are designed for SWIPT, PHY security, and multicasting, and have a more stable and 
	even better performance than existing methods. Higher secrecy rates are 
	achieved over MIMO wiretap channels with different 
	numbers of antennas. 
	Besides, it enlarges the data transmission rates for multicasting with lower 
	computational complexity. 
	\item Then, we propose a unified DNN-based  precoder which ``learns'' 
	from the conventional mathematical  models, including 
	analytical solutions 
	and RP. This precoding is able to 
	solve all objectives at the same time, since 
	the network  is trained by all the functions together.  To
	choose the objectives, we design an input feature utilizing binary code. The 
	performance of 
	the DNN-based 
	precoding is very close to that of the conventional 
	methods.
	\item The proposed DNN-based precoding is more efficient than the state-of-the-art iterative solutions. 
	Specifically, it is more than an order of magnitude faster 
	than numerical solutions 
	using a central 
	processing unit (CPU). In the scenario where a graphics 
	processing unit (GPU) is affordable (like base stations), it takes an average 
	execution time around
	$0.40$ms and $0.043$ms corresponding to 10 
	and 100 channels concurrently. 
	These numbers are much smaller than the coherence time of the wireless channels. 
	\item The performance of the  
	DNN-based 
	precoder is evaluated for different numbers of hidden layers 
	(\textit{depth}) and hidden nodes (\textit{width}). Increasing the depth and width positively increases the performance but increases the computational complexity.  
\end{itemize}

 \subsection{Organizations and Notations}\label{sec_intro_sub_org}
 The  remainder of this paper is organized as follows. We introduce the 
 system 
 models for the five objectives in Section~\ref{sec_sys}, and
 formulate rotation-based precoding in Section~\ref{sec_rot}.  In 
 Section~\ref{sec_net}, 
 we 
 propose the DNN architecture for our unified precoder. We illustrate  the training process and 
 results in Section~\ref{sec_data}. Finally, we conclude the 
 paper in Section~\ref{sec_conclu}.

 Notations: Bold lowercase letters denote column vectors and bold uppercase 
 letters denote matrices.  $ {a}_{i,j} $ represents the entry $ (i,j) 
 $ of matrix $ \mathbf{A} $. ${\rm 
 	vec}(\cdot)$ vectorizes a matrix by cascading columns.
 $|\cdot|$, 
 $(\cdot)^T$, 
 ${\rm ln}(\cdot)$, 
 ${\rm tr}(\cdot)$ are the absolute value, 
 Euclidean norm,  transpose, natural logarithm, respectively.  ${\rm 
diag}(\cdot)$  
 designates  the diagonal matrix of the set inside. ${\rm 
 	sign}(\cdot)$ extracts the sign of a real number. $E\{\cdot\}$ is the 
 expectation of random variables.  $[x]^+$ expresses the maximum value of 
 0 
 and $x$. $\mathbf{I}_a$ is an $a\times a$ identical 
matrix, and
 $\mathbf{0}_{a\times b}$ ($\mathbf{1}_{a\times b}$) is  
 all-zeros (all-ones) matrix of dimension 
 ${a\times b}$.

\section{System Models and Mathematical Preliminaries}\label{sec_sys}
\subsection{Channel Model}
\begin{figure}[t]
	\centering
	\includegraphics[width=0.68\textwidth]{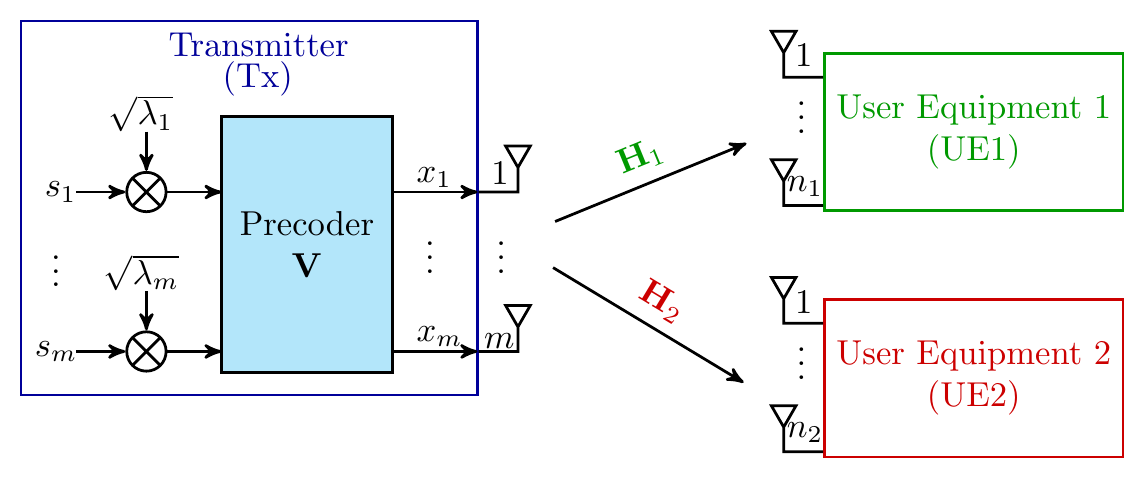}
	\caption{The two-user MIMO network with $m$, $n_1$, and $n_2$  
		antennas at Tx,  UE1, and  UE2, respectively. Here,  $\mathbf{x}\in\mathbb{R}^{m\times 1}$ is the transmitted signal, 	$\mathbf{V}\in\mathbb{R}^{m\times m}$ is the precoding matrix,
		and $\lambda_1,\hdots,\lambda_{m}$ 
		are the powers allocated to symbols $s_1,\hdots,s_{m}$, respectively.		 
		}
	\label{fig_figMIMOME}
\end{figure}

In this 
paper, we consider a   MIMO wireless communication system  
with one transmitter and two receivers.  The transmitter (Tx) is equipped with 
$m$ transmit
antennas and broadcasts information to the users. Inside Tx,
a linear precoder is applied as shown in Fig.~\ref{fig_figMIMOME}. In this 
figure,
$\mathbf{s}\triangleq[s_1,\hdots,s_{m}]^T$  is an independent and 
unit 
power symbol vector, that is, 
$\mathbb{E}\{\mathbf{s}\mathbf{s}^T\}=\mathbf{I}_m$.   
$\mathbf{\Lambda}\triangleq{\rm diag}(\lambda_1,\hdots,\lambda_{m})$ 
represents the power allocation matrix, and    
$\mathbf{V}\in\mathbb{R}^{m\times m}$ is the precoding matrix.  
Then,  
the transmitted signal $\mathbf{x}$ is 
\begin{align}
\mathbf{x} =\mathbf{V}\mathbf{\Lambda}^{\frac{1}{2}}\mathbf{s},
\end{align}
whose covariance matrix is 
$\mathbf{Q}\triangleq\mathbb{E}\{\mathbf{x}\mathbf{x}^T\}
=\mathbf{V}\mathbf{\Lambda}\mathbf{V}^T$.  
The channel input is
subject to an average total power constraint
\begin{align}
{\rm tr}(\mathbb{E}\{\mathbf{x}\mathbf{x}^T\}) \leq P. \label{eq_Pt}
\end{align}
At the receivers' side,  user equipment 1 (UE1)
and user equipment 2 (UE2) are equipped with  $n_1$ 	and $n_2$  
antennas, 
respectively. 
It is assumed that the transmission is over a flat fading channel. 
The input-output relations  are given as
\begin{subequations}
	\begin{align}
	\mathbf{y}_1 = \mathbf{H}_1\mathbf{x} + \mathbf{w}_1,\label{eq_recSig1}\\
	\mathbf{y}_2 = \mathbf{H}_2\mathbf{x} + \mathbf{w}_2,\label{eq_recSig2}
	\end{align}
\end{subequations}
in which $\mathbf{y}_1\in \mathbb{R}^{n_1\times 1}$ and $\mathbf{y}_2\in 
\mathbb{R}^{n_2\times 1}$ are received signals at UE1 and UE2,  
${\mathbf{H}_1} \in \mathbb{R}^{n_1 \times m}$  and $\mathbf{H}_2 
\in \mathbb{R}^{n_2 \times m}$ are the channels corresponding to  
UE1 and UE2,   and $\mathbf{w}_1 \in 
\mathbb{R}^{n_1\times 1}$ 
and $\mathbf{w}_2\in \mathbb{R}^{n_2\times 1}$ are independent and 
identically 
distributed (i.i.d)  Gaussian noises with zero means and identity covariance 
matrices. 
The above-mentioned MIMO communication system can have 
multiple objectives as described in the following. Throughout this paper, 
$\mathcal{O}_i$
refers to objective $i$, $i \in\{1,\dots,5\}$, as described in Table~\ref{tab_modes}.

%
%
%

\subsection{Objectives}

\subsubsection{WIT ($\mathcal{O}_1$)}\label{sec_sys_t1}
In this objective, UE1 acts as an information decoding  user who seeks for  
highest transmission rate over the MIMO channel, while UE2 is ignored.  
The information transmission capacity $\mathcal{C}_1$ is
obtained by solving the  following problem \cite{cover2012elements}  
\begin{subequations}
	\begin{align}\label{eq_opt1}
	\textmd{(P1)}
	\quad\mathcal{C}_1 \triangleq& \max\limits_{\mathbf{Q}}\frac{1}{2}\log 
	{|\mathbf{I}_{n_1}+\mathbf{H}_1\mathbf{Q}\mathbf{H}_1^T|},\\
	&{\;\;\rm s.t.\;}\mathbf{Q}\succeq\mathbf{0}, 
	\mathbf{Q}=\mathbf{Q}^T, 
	{\rm tr}(\mathbf{Q})\leq P.
	\end{align}
\end{subequations}
The optimal solution of 
(P1) is obtained  using singular value 
decomposition (SVD) and 
water-filling algorithm \cite{cover2012elements}. The optimal covariance 
matrix in (P1) can be 
expressed as 
\begin{align}\label{eq_p1Opt}
\mathbf{Q}_1^\ast=\mathbf{A}\mathbf{B}\mathbf{A}^T,
\end{align} 
in which $\mathbf{A}$ is obtained as the 
\textit{right-singular 
	vectors} of the 
channel
$\mathbf{H}_1$, and $\mathbf{B}$ in 
is obtained from water-filling algorithm \cite{cover2012elements}. 
Later, 
we will use this analytical solution to generate training sets
for  $\mathcal{O}_1$.

\subsubsection{EH ($\mathcal{O}_2$)}\label{sec_sys_t2}
EH refers to transmitting electrical energy originated from a power source. 
The transmitter emits radio-frequency signals, and UE2, as an EH user, tries 
to maximize the energy transmission efficiency. The objective function 
of this problem is shown in \cite{zhang2013mimo}
\begin{subequations}
	\begin{align}\label{eq_opt2}
	\textmd{(P2)}
	\quad\mathcal{C}_2 \triangleq& \max\limits_{\mathbf{Q}}\eta\cdot{\rm 
		tr}(\mathbf{H}_2\mathbf{Q}\mathbf{H}_2^T),\\
	&{\;\;\rm s.t.\;}\mathbf{Q}\succeq\mathbf{0}, 
	\mathbf{Q}=\mathbf{Q}^T, 
	{\rm tr}(\mathbf{Q})\leq P,
	\end{align}
\end{subequations}
where $\eta$ is the  \textit{converting rate} of  the harvested energy and, 
without loss of generality, 
we 
assume 
$\eta=1$ throughout the paper. The optimal analytical solution is given in 
\cite{zhang2013mimo}. It 
applies  SVD to decompose channel $\mathbf{H}_2$ as
$\mathbf{H}_2=\mathbf{E}\mathbf{F}\mathbf{G}^T$,
in which $\mathbf{E}$ and $\mathbf{G}$ are orthonormal matrices 
and 
$\mathbf{F}$ is a diagonal matrix that contains 
non-negative singular 
values. If the diagonal elements of $\mathbf{F}$ are in 
descending order,
then the 
optimal solution of (P2) is given as\cite{zhang2013mimo}
\begin{align}\label{eq_p2Opt}
\mathbf{Q}_2^\ast=P\mathbf{g}_1\mathbf{g}_1^T,
\end{align}
where 
$\mathbf{g}_1$ is  the first column of $\mathbf{G}$. 

\subsubsection{SWIPT ($\mathcal{O}_3$)}\label{sec_sys_t3-5}
SWIPT is to balance the WIT from Tx 
to UE1 and the EH at UE2 simultaneously. As defined 
in \cite{zhang2013mimo}, SWIPT characterizes the optimal trade-off between 
the maximum energy and information transfer by the \textit{rate-energy 
	region} which is formed as 
\cite{zhang2013mimo},
\begin{subequations}
	\begin{align}\label{eq_opt3} 
	\textmd{(P3)}
	\quad\mathcal{C}_3 \triangleq& 
	\max\limits_{\mathbf{Q}}\frac{1}{2}\cdot\log 
	{|\mathbf{I}_{n_1}
		+\mathbf{H}_1\mathbf{Q}\mathbf{H}_1^T|},\\
	&{\;\;\rm s.t.\;} \ 
	\eta\cdot{\rm tr}
	(\mathbf{H}_2\mathbf{Q}\mathbf{H}_2^T)\geq {\cal \bar E}, \\
	&\; \qquad \mathbf{Q}\succeq\mathbf{0}, \mathbf{Q}=\mathbf{Q}^T,
	{\rm tr}(\mathbf{Q})\leq P,
	\end{align}
\end{subequations}
in which $\cal\bar{E}$ is a dynamic threshold representing the  required 
minimum
energy harvested by UE2. The  value of $\cal\bar{E}$ is in
the range from minimum (${\cal 
	E}_{\min}$) to maximum (${\cal 
	E}_{\max}$),
\begin{align}\label{eq_thres}
{\cal{\bar{E}}}\triangleq{\cal E}_{\min} + q ({\cal E}_{\max}-{\cal E}_{\min}),
\end{align}
where $q$ is called the \textit{normalized EH level} and varies from $0\%$ 
to $100\%$. Since we are looking 
for the maximum 
rate-energy boundary, ${\cal E}_{\min}$ is defined as the energy received  
by UE2 when 
UE1 achieves the maximum data rate, i.e., (P1) reaches its optimal. Then, we 
have 
${\cal E}_{\min}=\eta\cdot{\rm 	
tr}(\mathbf{H}_2\mathbf{Q}_1^\ast\mathbf{H}_2^T)$
where $\mathbf{Q}_1^\ast$ is given in \eqref{eq_p1Opt}. 
On the other hand, 
${\cal E}_{\max}$ can be obtained when 
UE2 reaches the maximum EH level by solving (P2), i.e., ${\cal 
E}_{\max}=\mathcal{C}_2$.
When $q=0\%$ or $q=100\%$, (P3)   degenerates to (P1) and 
(P2), respectively.
%


\subsubsection{PHY Security ($\mathcal{O}_4$)}\label{sec_sys_t6}
Under this objective, UE1 is   a legitimate user  and requires 
services while keeping it secret from an eavesdropper, UE2. 
The precoder is expected to maximize the 
secrecy transmission  rate \cite{liu2009note}
\begin{subequations}
	\begin{align}\label{eq_opt4}
	\textmd{(P4)}
	\quad\mathcal{C}_4 \triangleq& \max\limits_{\mathbf{Q}}\frac{1}{2}\log 
	\frac{|\mathbf{I}_{n_1}+\mathbf{H}_1\mathbf{Q}\mathbf{H}_1^T|}{|\mathbf{I}_{n_2}
		+\mathbf{H}_2\mathbf{Q}\mathbf{H}_2^T|},\\
	&{\;\;\rm s.t.\;}\mathbf{Q}\succeq\mathbf{0},\mathbf{Q}=\mathbf{Q}^T, 
	{\rm tr}(\mathbf{Q})\leq P.
	\end{align}
\end{subequations}

An optimal analytical solution for MIMO wiretap channel only exist in limited 
cases, 
including the case when $n_t = 2$  \cite{vaezi2017journal}. 
Other solutions are  numerical  like  GSVD
\cite{fakoorian2012optimal}, AO-WF \cite{li2013transmit}, etc.

\subsubsection{Multicasting ($\mathcal{O}_5$)}\label{sec_sys_t7}
In this configuration, Tx offers a multicasting message to both 
users, 
such as
advertisements and emergency alerts. To ensure the multicasting message 
can be 
decoded by everyone,
multicasting rate is limited to the minimum
rate of the receivers. 
The transmission rate of this problem is 
formulated 
as \cite{zhu2012precoder}
\begin{subequations}
	\begin{align}\label{eq_opt5}
	\textmd{(P5)}
	\quad\mathcal{C}_5 \triangleq& \max\limits_{\mathbf{Q}}
	\; \min\limits_{u=1,2}
	\frac{1}{2}
	\log{|\mathbf{I}_{n_u}+\mathbf{H}_u\mathbf{Q}\mathbf{H}_u^T|}
	,\\
	&{\;\;\rm s.t.\;}\mathbf{Q}\succeq\mathbf{0},\mathbf{Q}=\mathbf{Q}^T, 
	{\rm tr}(\mathbf{Q})\leq P.
	\end{align}
\end{subequations}
In the MISO case,  SDR techniques yield a closed-form solution 
\cite{sidiropoulos2006transmit}. In the MIMO case, 
CAA is 
proposed in 
\cite{zhu2012precoder}, and
\cite{tan2015opportunistic} mentions that the 
problem can be 
solved by semidefinite programming (SDP) directly. 
Moreover,  \cite{vaezi2019rotation} 
introduced a nonlinear random search with rotation parameters. 
However, 
the
existing methods are limited to the computational complexity
or are only available for a specific number of antennas.

As we saw, (P1)-(P5) have different expressions and solutions in general. 
As the objective changes, the corresponding solution changes completely. 
Moreover, (P3)-(P5) can only be solved iteratively which incur high complexity and thus are slow. To tackle this, we first 
propose a  
unified 
solution for (P3)-(P5), which is robust and reliable in a 
variety of antenna settings. Then, we propose DNN-based precoding that 
can solve (P1)-(P5) simultaneously and efficiently.

\section{Rotation-based Precoding}\label{sec_rot}
In this section, we introduce RP and apply it to (P3)-(P5). We should highlight that RP can also be applied to (P1)-(P2) but these two problems have competitive analytical solutions, and there is no need to a new solution.

\subsection{Rotation-based Precoding (RP)}
The covariance matrix $\mathbf{Q}$  can be formed using eigenvalue 
decomposition as
\begin{align}\label{eq_eig1}
\mathbf{Q}\triangleq \mathbf{V} \mathbf{\Lambda} \mathbf{V}^T,
\end{align}
in which $\mathbf{\Lambda}\in\mathbb{R}^{m\times m}$ is a diagonal 
matrix, 
whose diagonal elements $[\lambda_1,\hdots,\lambda_{m}]$ are 
non-negative due to the PSD constraint.
Then, the average power constraints in (P1)-(P5) are equivalent to 	
$\sum_{i=1}^{m}\lambda_i\leq P$. Thus, the PSD and power constraints 
can be represented as a set of  linear constraints
\begin{align}\label{eq_allcon1}
   \{\lambda_{i}| \lambda_{i}\geq0, \; \sum_{i=1}^{m}\lambda_i\leq P\}.
\end{align}
Besides, $ \mathbf{V}\in\mathbb{R}^{m\times m} $ is an orthonormal matrix 
due to the symmetric property of $\mathbf{Q}$.  
It can be modeled as a Given's matrix 
\cite{matrix,zhang2019rotation} 
also 
named as a rotation matrix
\begin{align}\label{eq_Vnbyn_}
\mathbf{V}=\prod_{i=1}^{m-1}\prod_{j=i+1}^{m} \mathbf{V}_{i,j},
\end{align}
where $ \mathbf{V}_{i,j} $ is  an identity matrix except for four elements
\begin{align}\label{eq_VnDsub}
\left[
\begin{matrix} 
v_{i,i}	&v_{i,j}\\
v_{j,i}	&v_{j,j}
\end{matrix}
\right]
=\left[
\begin{matrix}
\cos\theta_{i,j}	&-\sin \theta_{i,j}\\
\sin\theta_{i,j}	&\cos \theta_{i,j}
\end{matrix}
\right].
\end{align}
Intuitively, for any vector 
$\mathbf{v}$ in $\mathbb{R}^{m\times1}$ vector space,  
$\mathbf{V}_{i,j}\cdot\mathbf{v}$ represents a rotating  from the 
$i$th standard basis 
to the $j$th standard basis with a certain rotation angle $\theta_{i,j}$. 
In total, we need\footnote{Specially, for $m=1$, $\mathbf{Q}$ becomes a 
scalar. In RP, we only have one  eigenvalue and no 
rotation angles. In such a case, \eqref{eq_numAngle} becomes $0$.}
\begin{align}\label{eq_numAngle}
n_a=\frac{1}{2}m(m-1),
\end{align}
rotation angles to represent $\mathbf{V}$ in \eqref{eq_Vnbyn_}. There 
is no  constraint on rotation angles, i.e., $\theta_{i,j}\in\mathbb{R}$. 	
In \cite{zhang2019rotation}, we have proved that an arbitrary covariance 
matrix $\mathbf{Q}$ can be represented by $m$ non-negative eigenvalues 
and $n_a$ rotation angles. Therefore, the optimization on $\mathbf{Q}$ 
can be equivalently transformed to optimization parameters using RP with 
the  constraint \eqref{eq_allcon1}.

It is worth mentioning that the order of multiplication in \eqref{eq_Vnbyn_} 
is not unique and different order will lead to different rotation angles 
$\theta_{i,j}$. 
In this paper, without loss of generality, we use the order defined in 
\eqref{eq_Vnbyn_}. Then, the rotation parameter vector can be defined as
\begin{align}\label{eq_allcon2}
\mathbf{r}\triangleq[{\bm\lambda},{\bm\theta}]^T,
\end{align}
where
	\begin{align}
	{\bm\lambda}\triangleq[\lambda_{1},\hdots,\lambda_{m}] \; \text{and} \;	
	{\bm\theta}\triangleq[\theta_{1,2},\hdots,\theta_{m-1,m}].
	\end{align}
	To this end, $\mathbf{Q}$ can be specified 
by the 
parameter vector $\mathbf{r}$ with the new
constraint
\begin{align}
\mathbf{L}\mathbf{r}\leq\mathbf{b},\label{eq_linear_const}
\end{align}
 where
\begin{equation}\textbf{}
\mathbf{L}\triangleq\left[
\begin{matrix}
-\mathbf{I}_m & \mathbf{0}_{m\times n_a}\\
\mathbf{1}_{1\times m} & \mathbf{0}_{1\times n_a}
\end{matrix}
\right]
\textrm{and} \;
\mathbf{b}\triangleq\left[
\begin{matrix} 
\mathbf{0}_{1\times m}\\
P
\end{matrix}
\right].
\end{equation}

\subsection{Rotation-based Precoder for $\mathcal{O}_3$ to $\mathcal{O}_5$ }
The problems \textmd{(P3)}-\textmd{(P5)} are challenging
and optimal analytical precoding matrices are not 
 known. In the following, we apply RP to the 
problems, which can parameterize all of the problems with rotation  angles 
and power allocation parameters. 
\subsubsection{RP for SWIPT}\label{sec_rot_t3-5}
Applying the RP on (P3), the objective function of 
SWIPT becomes
\begin{subequations}
	\begin{align}\label{eq_opt3a} 
	\textmd{(P3a)}
	\quad\mathcal{C}_3 =
	& \max\limits_{\mathbf{r}}\frac{1}{2}\log 
	{|\mathbf{I}_{n_1}
		+\mathbf{H}_1\mathbf{Q}\mathbf{H}_1^T|},\\
	&{\;\;\rm s.t.\;}\mathbf{L}\mathbf{r}\leq\mathbf{b},\label{eq_P3a_c1}\\
	&\qquad\ 
	\eta\cdot{\rm tr}
	(\mathbf{H}_2\mathbf{Q}\mathbf{H}_2^T)\geq {\cal \bar 
	E}.\label{eq_P3a_c2}
	\end{align}
\end{subequations}
This problem can be solved by a general optimization tool such as 
\texttt{fmincon} in 
\textsc{Matlab}. Here, \eqref{eq_P3a_c1} is a linear inequality 
constraint  and \eqref{eq_P3a_c2} can 
be added as non-linear constrain in \texttt{fmincon}. The 
For initialization of $\mathbf{Q}$ we use the solution of (P2), i.e., 
$\mathbf{Q}_2^\ast$ in \eqref{eq_p2Opt}. 
Then, we obtain 
the initial value $\mathbf{r}$ using \eqref{eq_eig1}-\eqref{eq_VnDsub} or 
Algorithm~1 in 
\cite{zhang2019rotation}. Finally, we can obtain $\mathbf{Q}_3^\ast$ 
which is defined to be the optimal solution for 
$\mathcal{O}_3$.

\subsubsection{RP for PHY Security}\label{sec_rot_t6}
	Applying RP to PHY Security problem results in 
\begin{subequations}
	\begin{align}\label{eq_opt4a}
	\textmd{(P4a)}
	\quad\mathcal{C}_4 =
	& \max\limits_{\mathbf{r}}\frac{1}{2}\log 
	\frac{|\mathbf{I}_{n_1}+\mathbf{H}_1\mathbf{Q}\mathbf{H}_1^T|}{|\mathbf{I}_{n_2}
		+\mathbf{H}_2\mathbf{Q}\mathbf{H}_2^T|},\\
	&{\;\;\rm s.t.\;}\mathbf{L}\mathbf{r}\leq\mathbf{b}.
	\end{align}
\end{subequations}	
Then, this new optimization problem can be solved by convex toolbox 
such as \texttt{fmincon} in \textsc{Matlab}. Then, we can obtain 
$\mathbf{Q}_4^\ast$ for 
	$\mathcal{O}_4$. Although the PHY security is  known 
as a non-convex problem, the performance of RP is more  
reliable compared with 
existing solutions,  {such as GSVD
\cite{fakoorian2012optimal} and AO-WF \cite{li2013transmit}.}

\subsubsection{RP for Multicasting} \label{sec_rot_t7}
Similarly, (P5) can be reformed as
	\begin{subequations}\label{eq_opt5a}
	\begin{align}
	\textmd{(P5a)}
	\quad\mathcal{C}_5 =& \max\limits_{\mathbf{r}}
	\; \min\left\{
	R_u
	\right\},u=1,2,\\
	&{\;\;\rm s.t.\;}\mathbf{L}\mathbf{r}\leq\mathbf{b},\label{eq_P5a_const}
	\end{align}
\end{subequations}
where $R_u$  represents the WIT rates of UE1 and UE2, i.e., 
\begin{subequations}
	\begin{align} \label{multicast}
	R_u(\mathbf{Q}) \triangleq 
	\frac{1}{2}\log{|\mathbf{I}_{n_u}+\mathbf{H}_u\mathbf{Q}\mathbf{H}_u^T|}, 
	u=1,2.
	\end{align}
\end{subequations}
{(P5a) is the minimum of two WIT problems 
represented in
	(P1) which is concave \cite{cover2012elements}. Thus, (P5a) is 
	concave. 
	Define the optimal solutions $\mathbf{Q}_{1}^{*(1)}$ and 
	$\mathbf{Q}_{1}^{*(2)}$  for $R_{1}$  and  $R_{2} $ in \eqref{multicast}.
 Then, the (P5a) in \eqref{eq_opt5a} can be solved by three {sub-cases}:
	\begin{itemize}
	\item \textit{Case 1:}  $ R_{1}(\mathbf{Q}_{1}^{*(1)})\leq 
	R_{2}(\mathbf{Q}_{1}^{*(1)})$, then 
	the optimal 
	multicast covariance matrix of \eqref{eq_opt5a} is $\mathbf{Q}^{*}_5= 
	\mathbf{Q}_{1}^{*(1)}$. 
	\item \textit{Case 2:}  $ R_{1}(\mathbf{Q}_{1}^{*(2)})\geq 
	R_{2}(\mathbf{Q}_{1}^{*(2)})$, 	the optimal 
	multicast covariance matrix of \eqref{eq_opt5a} is $\mathbf{Q}^{*}_5= 
	\mathbf{Q}_{1}^{*(2)}$. 
	\item \textit{Case 3:} Otherwise, we solve the rotation parameters in 
	\eqref{eq_opt5a} using 	\texttt{fmincon}. 
\end{itemize}
}
	Since the first two sub-cases are actually  WIT problems with analytical 
solutions, the efficiency of the solution improves
compared to 
iterative solutions such as CAA \cite{zhu2012precoder} and SDP.
 Till now, the solutions $\mathbf{Q}^{*}_3$ to $\mathbf{Q}^{*}_5$ for 
$\mathcal{O}_3$ to $\mathcal{O}_5$ is obtained 
using the RP, respectively. 
In the next section, we propose using supervised DNN  
to learns from the above solutions and find the covariance matrices 
corresponding to (P1)-(P5) at once.


\section{A Unified DNN-based Precoder}\label{sec_net}
In this section, we introduce  
a unified DNN-based precoding and power allocation, including the DNN structure, the input features, and 
network outputs. DNN can increase the efficiency by unifying the solution for all of 
the problems together in contrast to the conventional methods which perform optimization  one by one. 

Before talking about the details of the DNN, we indicate that $\mathcal{O}_3$ (the SWIPT problem) will be differentiated for nine 
normalized EH level in 
\eqref{eq_thres} for
\begin{align}\label{eq_qvec}
	q\in\{90\%, 80\%,\hdots,10\%\}.
\end{align}
These sub-problems are named $\mathcal{O}_3$(90$ \% $)    to    
$\mathcal{O}_3$(10$ \% $)  as shown in Table~\ref{tab_code}. Then, for 
the proposed DNN, a unique index $\mathcal{M}_k$, 
$k\in\{1,\hdots,K\}$, represents different modes. 
Therefore, we have $K=13$ modes in total.
It is worth clarifying that  $\mathcal{O}_i$ refers to 
the configurations of  precoding objectives, i.e., WIT, EH, etc.,  while we use  
 $\mathcal{M}_k$ for integer index and  simplifying the expressions. These are listed in Table~\ref{tab_code} for clarity.

\subsection{The DNN Structure}
\begin{figure}[h]
	\centering
	\includegraphics[width=0.6\textwidth]{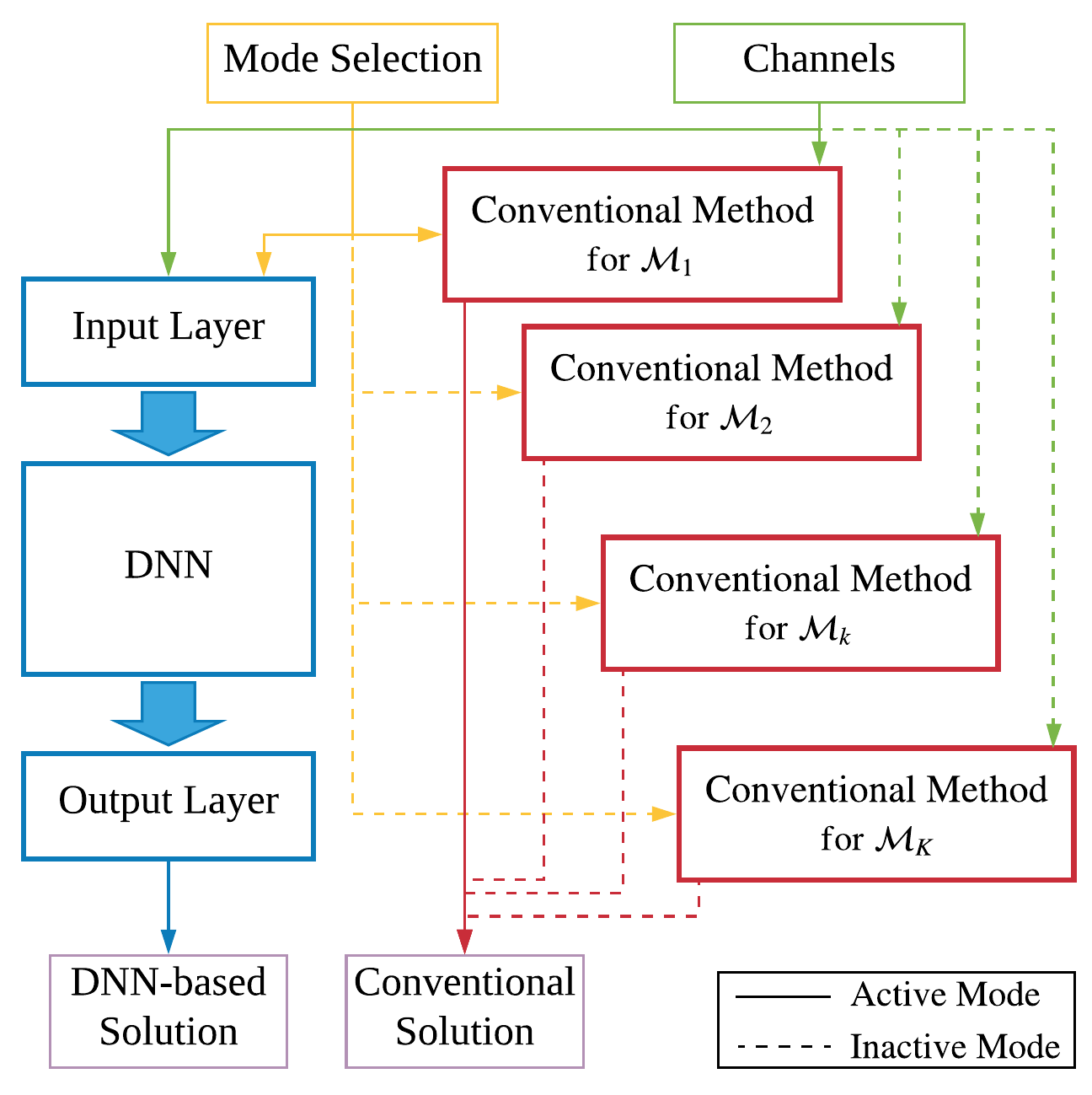}
	\caption{The structure of the proposed DNN-based precoding. 
		The current selected mode is  $\mathcal{M}_k$.}
	\label{fig_train}
\end{figure}

The structure of the proposed DNN-based precoding is demonstrated in
Fig.~\ref{fig_train}. At the top, the parameters of the two users, including 
mode and channel selection. The DNN-based precoder can provide the 
precoding solution 
directly, while it is necessary to activate one of the conventional methods 
according  to the user requirement. 
The DNN-based precoder can be divided into three parts shown by different 
colors in Fig.~\ref{fig_train}. These are
1) the input layer, which pre-processes the input information and 
generates a feature vector for DNN; 2) DNN is applied to achieve the 
non-linear mapping between the features and demanded outputs; 3) the 
output layer that maps the output to a covariance matrix for 
precoding. 

\begin{figure}[t]
	\centering
	\includegraphics[width=0.65\textwidth]{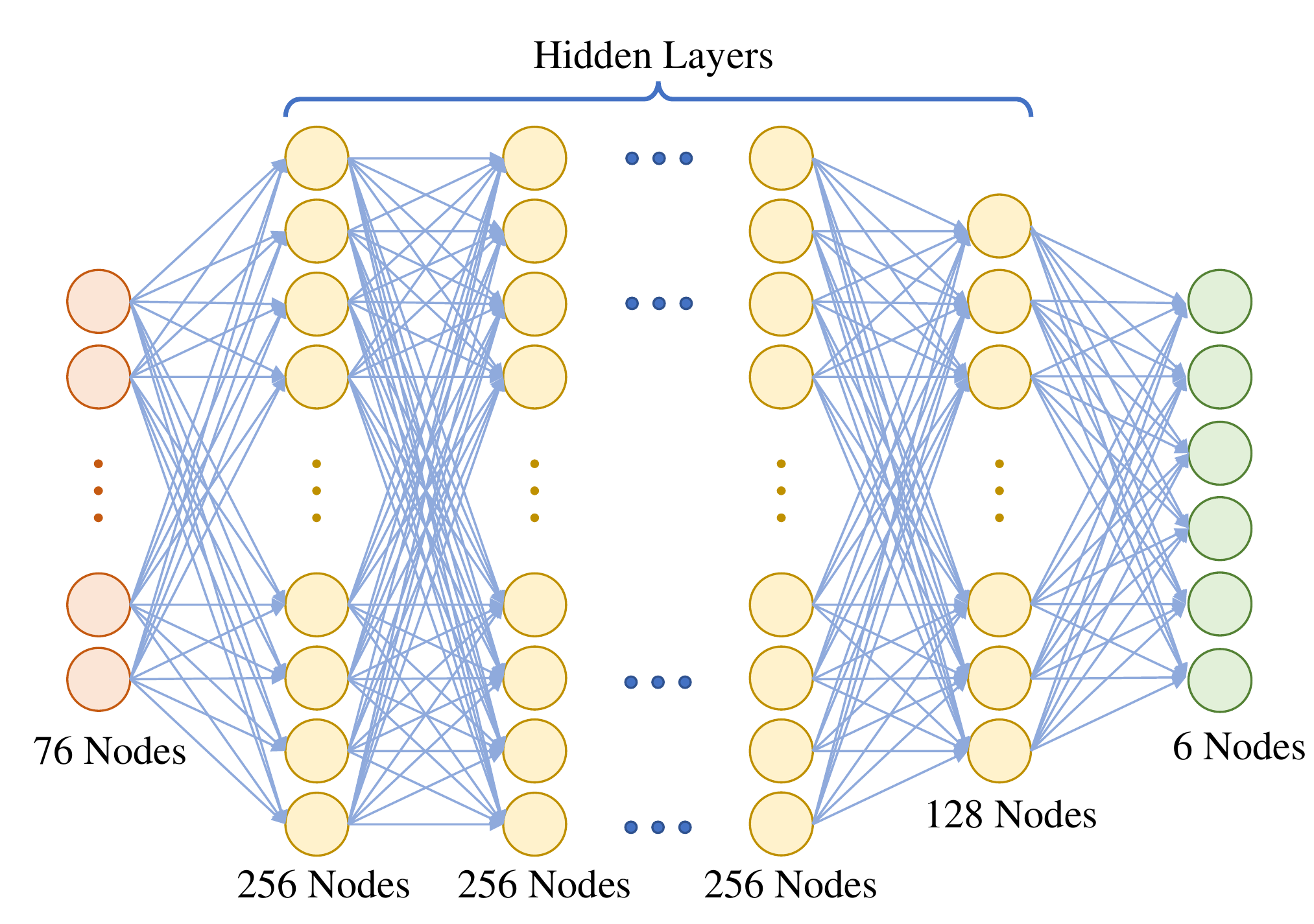}
	\caption{The DNN architecture.}
	\label{fig_network}
\end{figure}
\begin{figure}[t]
	\centering
	\includegraphics[width=0.658\textwidth]{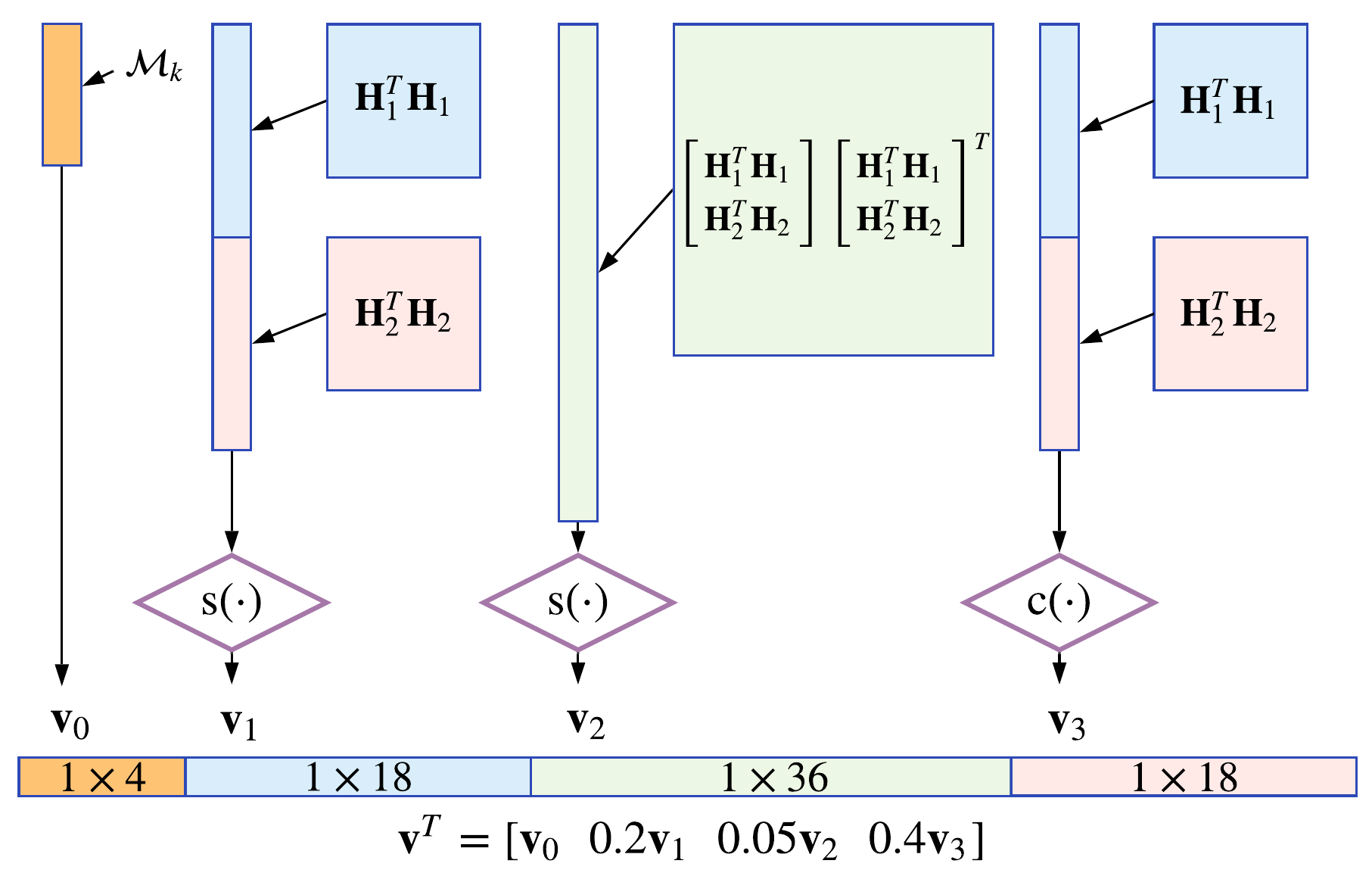}
	\caption{Feature design for  the input layer.}
	\label{fig_input}
\end{figure}
As shown in Fig.~\ref{fig_network}, the  architecture of the DNN 
{precoder}
has ten  fully-connected  hidden layers 
equipped with \textit{parametric 
	rectified linear units} (PReLU) 
\cite{he2015delving} as activation functions. 
PReLU is defined as 
\begin{align}
f(y)=\left\{\begin{matrix}
y, &y\geq0,\\
\alpha y, &y<0,\\
\end{matrix}\right.
\end{align}
where $\alpha=0.25$ is 
	a trainable initialized parameter\footnote{We have 
		examined the performance of PReLU initialized with a fixed value 0.25 given 
		in \cite{he2015delving} and random uniformly distributed values. The 
		fixed 
		initialization achieves better performance, especially for
		$\mathcal{O}_4$.}. 
	PReLU  
extends 
the freedom of the DNN to mimic a mapping and preventing over-fitting at 
the same time. 

\subsection{Input Features}

\begin{table}[t]
	\caption{Code Vector Design for Each Mode}
	\label{tab_code}
	\centering
	\begin{tabular}{llc|llc}
		\hline
		 Objective &  Mode     & Code ($\mathbf{c}_k$)     &  Objective   & 
		 Mode & 
		Code  ($\mathbf{c}_k$)      \\ \hline
		$\mathcal{O}_1$ &$\mathcal{M}_1$    &  0001  & 	
		$\mathcal{O}_3${$(40\%)$}   & $\mathcal{M}_8$ & 
		1000 \\
		$\mathcal{O}_2$          &$\mathcal{M}_2$    & 
		0010& 		$\mathcal{O}_3${$(30\%)$}   & $\mathcal{M}_9$ & 
		1001 \\
		$\mathcal{O}_3${$(90\%)$}      &$\mathcal{M}_3$    &  
		0011 & 	$\mathcal{O}_3${$(20\%)$}   & 
		$\mathcal{M}_{10}$ 
		& 
		1010 \\
			$\mathcal{O}_3${$(80\%)$}       &$\mathcal{M}_4$    & 
			0100 & 	$\mathcal{O}_3${$(10\%)$}   & 
			$\mathcal{M}_{11}$ 
		& 
		1011 \\
		$\mathcal{O}_3${$(70\%)$}       &$\mathcal{M}_5$    &  
		0101  & 	$\mathcal{O}_4$   & $\mathcal{M}_{12}$   
		& 
		1100 \\
			$\mathcal{O}_3${$(60\%)$}       & $\mathcal{M}_6$    
			& 
			0110  & $\mathcal{O}_5$ & $\mathcal{M}_{13}$    
		& 
		1101 \\
			$\mathcal{O}_3${$(50\%)$}      &$\mathcal{M}_7$    &  
			0111 & \textit{}   & \textit{}      
		&   \\      
		\hline
	\end{tabular}
\end{table}

The schematic diagram of the input layer is shown in Fig.~\ref{fig_input}. The 
required  inputs  are $\mathbf{H}_1$, $\mathbf{H}_2$, and mode index, 
which given by  $k\in\{1,\hdots,K\}$. The feature vector contains four 
sub-features, namely, $\mathbf{v}_0$, $\mathbf{v}_1$, 
$\mathbf{v}_2$, and $\mathbf{v}_3$ which are 
\begin{subequations}
	\begin{align}
	&\mathbf{v}_0\triangleq    \mathbf{c}_k,  %
	\label{eq_inputV0}\\
	&\mathbf{v}_1\triangleq       {\rm s}({\rm vec}(\mathbf{F}))^T, 
	\label{eq_inputV1}\\
	&\mathbf{v}_2\triangleq  {\rm s}(   {\rm 
	vec}(\mathbf{F}^T\mathbf{F}))^T, 
	\label{eq_inputV2}\\
	&\mathbf{v}_3\triangleq {\rm c}({\rm vec}(\mathbf{F}))^T, 
	\label{eq_inputV3}
	\end{align}
\end{subequations}
where $\mathbf{c}_k\in\mathbb{R}^{1\times 4}$ is a  
\textit{binary-code vector} identifying the $k$-th objective. The objectives and 
the	corresponding code  are listed in Table~\ref{tab_code}. 
$\mathbf{F}$ contains channel information defined as
\begin{align}\label{eq_feature0}
\mathbf{F}\triangleq[\mathbf{H}_1^T\mathbf{H}_1 \;\;
\mathbf{H}_2^T\mathbf{H}_2].
\end{align}
${{\rm s}(x)}$ is the element-wise square root function keeping the sign of 
input $x$, i.e., 
\begin{align}\label{eq_sqrt}
{\rm s}(x)\triangleq {\rm sign}(x)\cdot|x|^{\frac{1}{2}}, 
\end{align}
Similarly, ${\rm c}(x)$ is the element-wise  cubic 
root function, which defined as 
\begin{align}\label{eq_cbrt}
{\rm c}(x)\triangleq {\rm sign}(x)\cdot|x|^{\frac{1}{3}}.
\end{align}

Among the sub-feature vectors, $\mathbf{v}_0$ represents the feature with 
respect to the mode index. With such a definition, the DNN has a better 
performance in recognizing the input objectives. Besides, 
$\mathbf{v}_1$, $\mathbf{v}_2$, and $\mathbf{v}_3$ are formed based on 
channels matrices. We use $\mathbf{H}_1^T\mathbf{H}_1$ 
and $\mathbf{H}_2^T\mathbf{H}_2$ rather than using $\mathbf{H}_1$ and 
$\mathbf{H}_2$ directly,
since 
\begin{subequations}
\begin{align}
&|\mathbf{I}+\mathbf{H}\mathbf{Q}\mathbf{H}|=|\mathbf{I}+\mathbf{H}^T\mathbf{H}\mathbf{Q}|,\\
&{\rm tr}(\mathbf{H}\mathbf{Q}\mathbf{H}^T)={\rm 
tr}(\mathbf{H}^T\mathbf{H}\mathbf{Q}),
\end{align}
\end{subequations}
can be applied to  optimization problems introduced in Section~\ref{sec_sys}.
 With such a definition, the dimension of input  features is related 
 only to $m$ rather than $m$, $n_1$, and $n_2$. Furthermore, non-linear 
 combinations  of channels are also considered  to deliver more information 
 to the DNN  in order to achieve a better non-linear {capability}.



Considering that the distribution of the channel elements are  
Gaussian,  the elements of $\mathbf{F}$  have a high density  
concentrating
around $0$. This 
 decreases the  fairness and distinguishability of the input features.
To overcome this  problem, we  use square root and 
cubic 
root in \eqref{eq_inputV1}-\eqref{eq_inputV3} to flattens  the 
distribution of the features to some degree. Finally, the input feature vector  $\mathbf{v}$ is a 
cascade of these sub-feature  vectors
\begin{align}\label{eq_inputVec}
\mathbf{v}\triangleq[\mathbf{v}_0, 0.2\mathbf{v}_1, 
0.05\mathbf{v}_2,0.4\mathbf{v}_3]^T,
\end{align}
where the coefficients  are chosen experimentally to normalize 
the sub-feature vectors and improve the accuracy of the DNN and the speed 
of learning  \cite{sola1997importance,lecun2012efficient}.
In this paper, we consider $P=20$ and $m=3$ while $n_1$ and $n_2$ can 
be any number based on specific cases. In this scenario, the size of the  input 
of the DNN is $76$ {\footnote{The input vector $\mathbf{v}$ contains $27$ 
pairs  of the same features since $\mathbf{H}_1^T\mathbf{H}_1$ and 
$\mathbf{H}_2^T\mathbf{H}_2$ are symmetric. Such  redundancy will be 
automatically reduced by the first hidden layer of  the DNN  
\cite{lecun2012efficient}.
}.

%
\subsection{Network Outputs}
Since the covariance matrix  is symmetric, the output (vector 
$\mathbf{q}$)  contains 
only upper triangular elements of $\mathbf{Q}$. That is, 
\begin{align}\label{eq_onputVec}
\mathbf{q}\triangleq[q_{1,1},q_{2,2},q_{3,3},q_{1,2},q_{2,3},q_{1,3}]^T.
\end{align}
Then, $ \mathbf{Q}$ can be assembled as
\begin{align}
\label{eq_Qmat}
\mathbf{Q}=\left[
\begin{matrix}
q_{1,1} & q_{1,2} & q_{1,3}\\
q_{2,1} & q_{2,2} & q_{2,3}\\
q_{3,1} & q_{3,2} & q_{3,3}
\end{matrix}
\right]= \mathbf{V} \mathbf{\Lambda} \mathbf{V}^T.
\end{align}
The precoding and power allocation matrices $\mathbf{V}$ and
$\mathbf{\Lambda}$, respectively, are obtained by eigenvalue 
decomposition. To ensure the PSD and total power constraints, negative 
diagonal elements in $\mathbf{\Lambda}$ are normalized to zero and the 
trace is scaled to $P$.
In the training procedure, $\mathbf{q}$ is known through RP  
corresponding to the problem discussed in 
Section~\ref{sec_rot}. Whereas, during testing,  $\mathbf{Q}$ 
will be 
obtained from the output vector and the precoding solution is obtained by 
eigenvalue decomposition \cite{zhang2019deep}. 
\section{Training procedure and Numerical Results}\label{sec_data}
In this section, we initially verify the performance of  RP 
which is used to train the network. Then, we explain 
the details of our data set and the  training procedure. Finally, 
we examine 
the performance of
the proposed DNN-based precoding.
\subsection{Performance of RP}
\subsubsection{SWIPT ($\mathcal{O}_3$)}
In Fig.~\ref{fig_SWIPTcomp}, the {performance of} RP is compared with the
\textit{time-switching and power-splitting} (TS-PS) \cite{zhang2013mimo} 
and random trials of $\mathbf{Q}$. For RP and TS-PS,  eleven thresholds 
${\cal{\bar{E}}}$  equally dividing the interval $[{\cal E}_{\min},{\cal 
	E}_{\max}]$ 
are considered in the cases of $P=10$, $20$, and $40$ (W).
The channels matrices, which were generated randomly, are
\begin{subequations}\label{eq_ch_sw}
	\begin{align}
	&\mathbf{H}_1
	= \left[\begin{matrix}
-2.2975& 0.4896&-1.8310\\
1.4576&-0.6100& 0.3800\\
0.8998& 0.0916&-0.3128
	\end{matrix}\right],\\
	&\mathbf{H}_2
	= \left[\begin{matrix}
-0.3276 &   3.3159 &  -0.9956\\
1.5765  &  0.2604  &  0.2578\\
-0.3337  &  1.1478 &  -0.3364
	\end{matrix}\right].
	\end{align}
\end{subequations}
For both channels, RP  can reach the same rate-energy region as 
TS-PS. The random trials are based on $10,000$  realizations of 
$\mathbf{Q}$. 
\begin{figure}[t]
	\centering
	\includegraphics[width=0.65\textwidth]{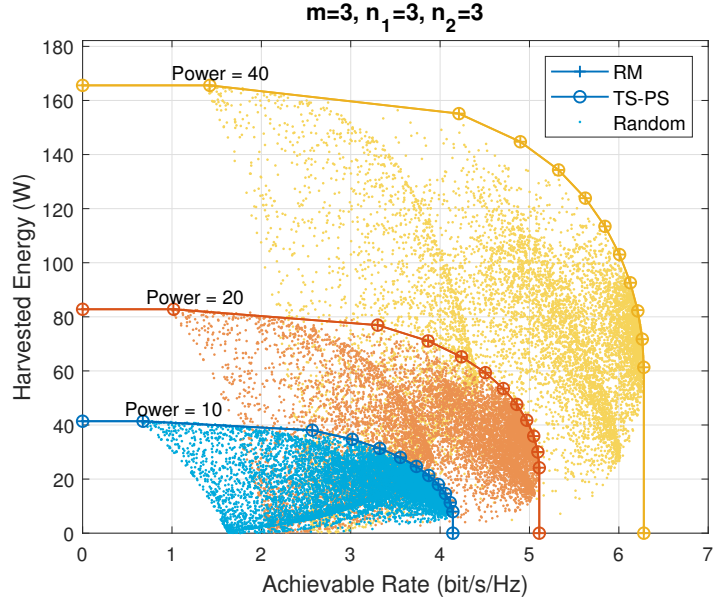}
	\caption{Comparisons of SWIPT region for the 
		RP and TS-PS for 10,000 random trials. 
		The colors distinguish different  transmit powers.}
	\label{fig_SWIPTcomp}
\end{figure}
\begin{figure*}[t]
	\centering
	\subfigure[Relative improvement ($\%$) of RP to GSVD.]{
		\includegraphics[width=0.31\textwidth]{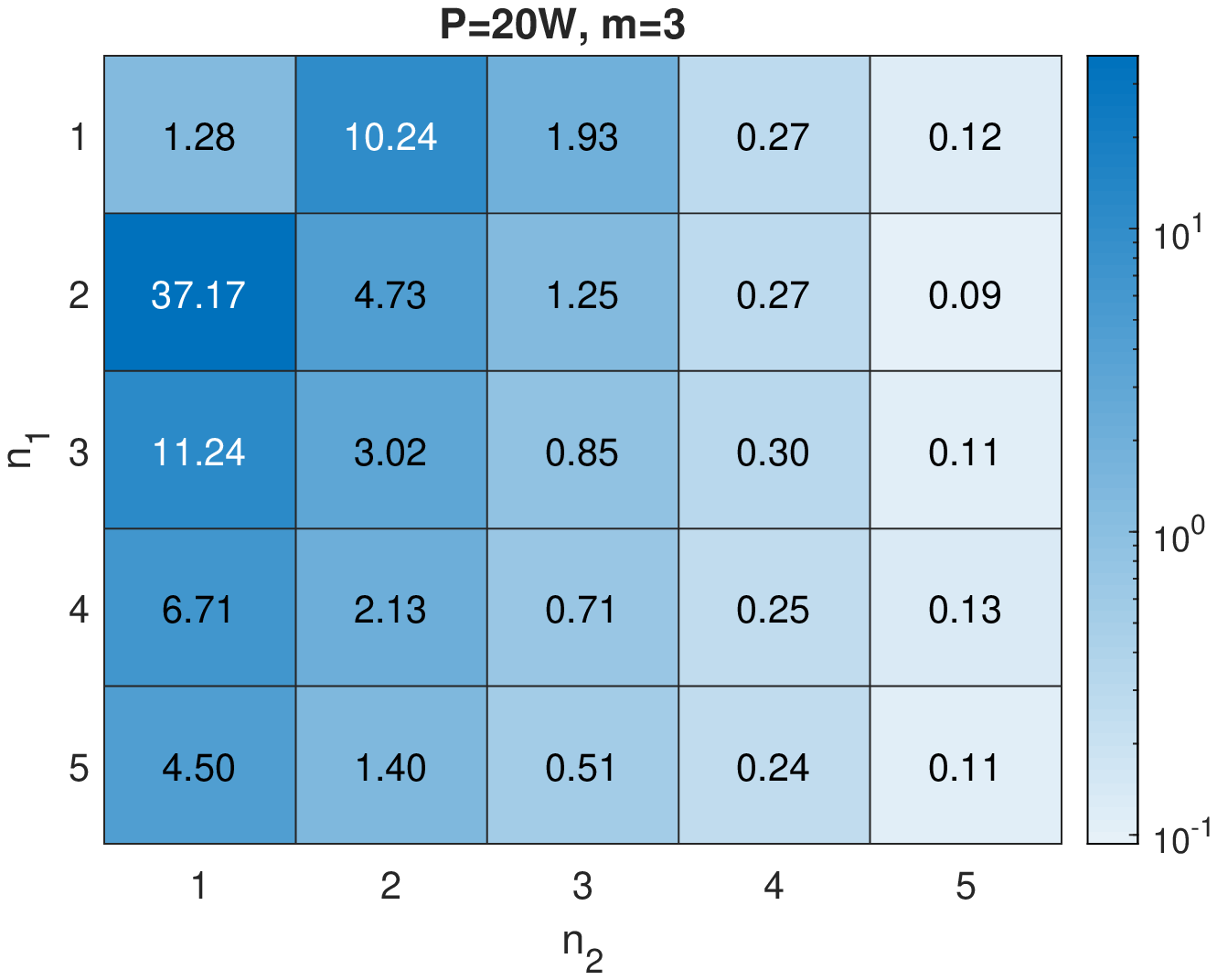}\label{fig_WTcomp_a}}
	\subfigure[Relative improvement ($\%$) of RP to AO-WF.]{
		\includegraphics[width=0.31\textwidth]{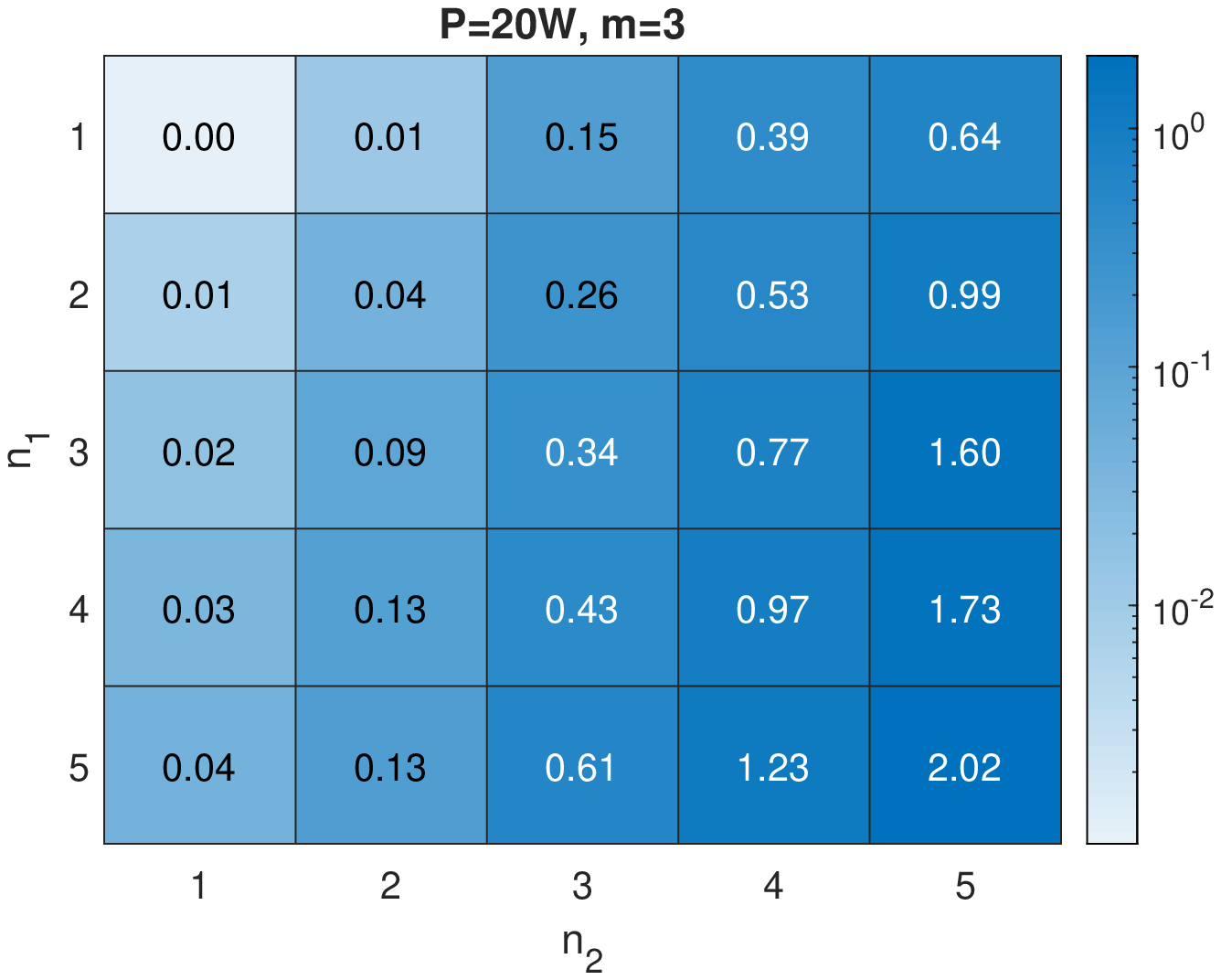}\label{fig_WTcomp_b}}
	\subfigure[Typical cases that  RP exceed  GSVD or 
	AO-WF.]{
		\includegraphics[width=0.33\textwidth]{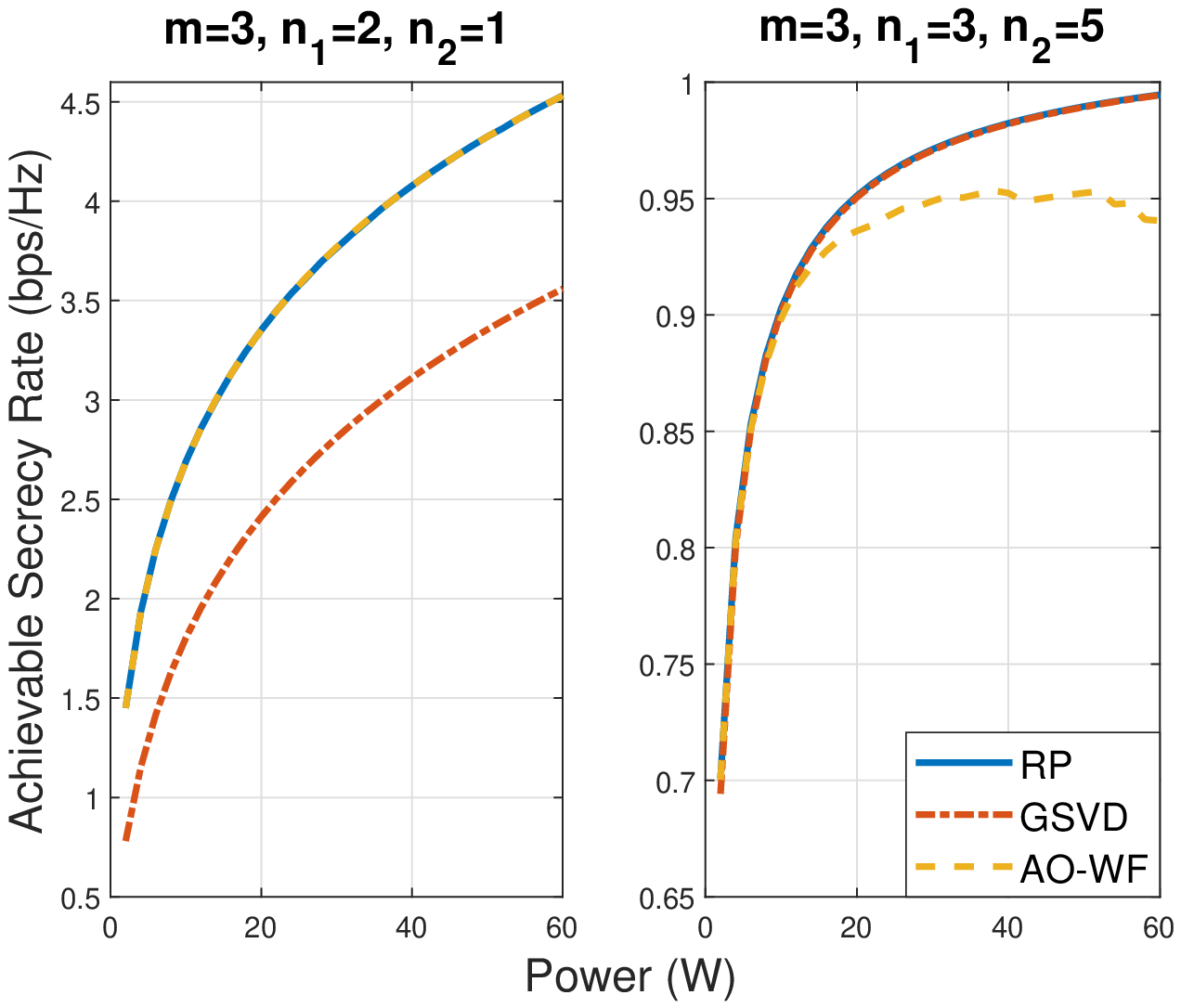}\label{fig_WTcomp_c}}
	\caption{Comparisons between the secrecy transmission rate of the 
		RP with GSVD and AO-WF.}
	\label{fig_WTcomp}
\end{figure*}
\begin{figure*}[h]
	\centering
	\subfigure
	[Relative improvement ($\%$) of RP to CAA.]{
		\includegraphics[width=0.31\textwidth]{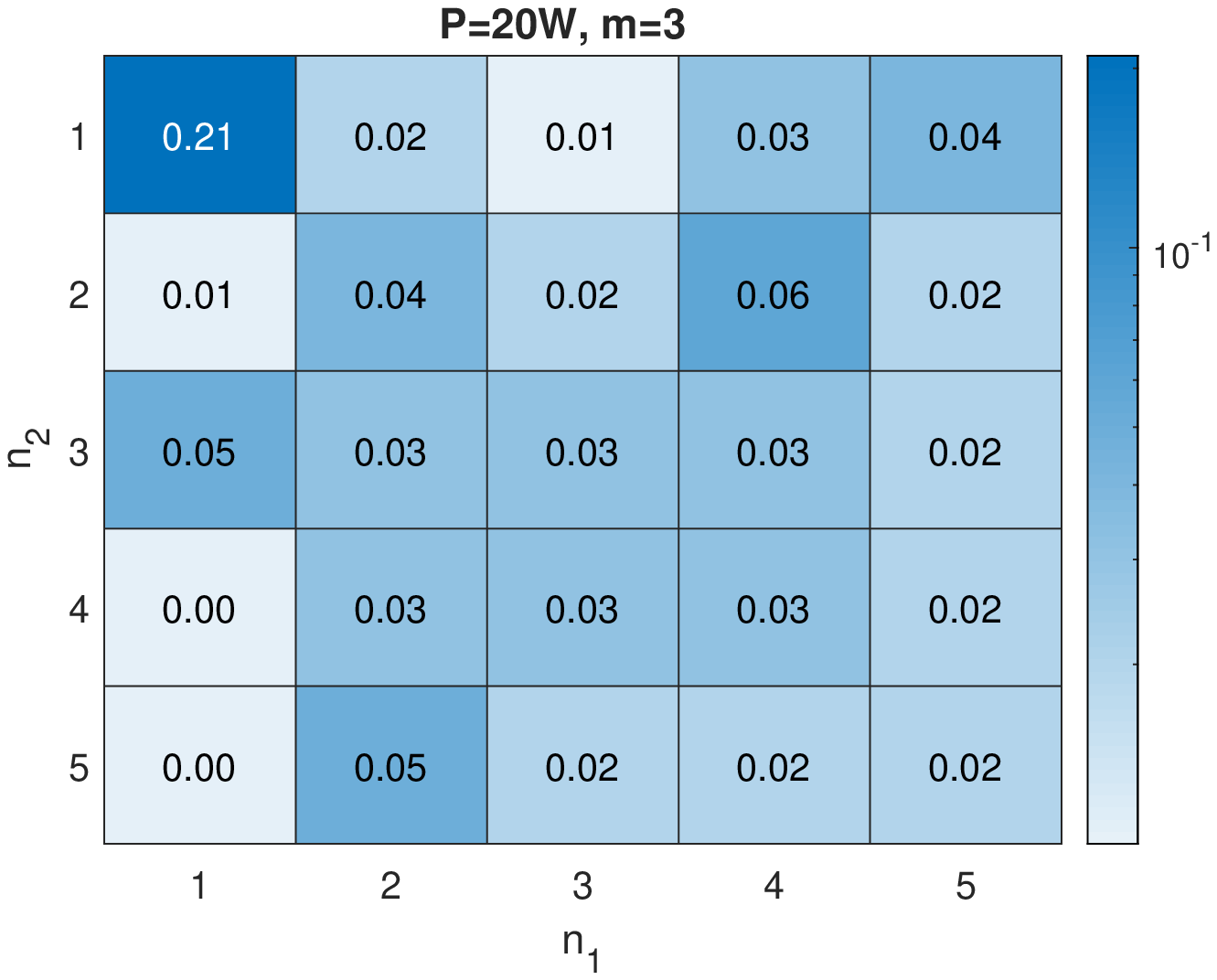}\label{fig_BCmap1}}
	\subfigure
	[Relative improvement ($\%$) of RP to SDP.]{
		\includegraphics[width=0.31\textwidth]{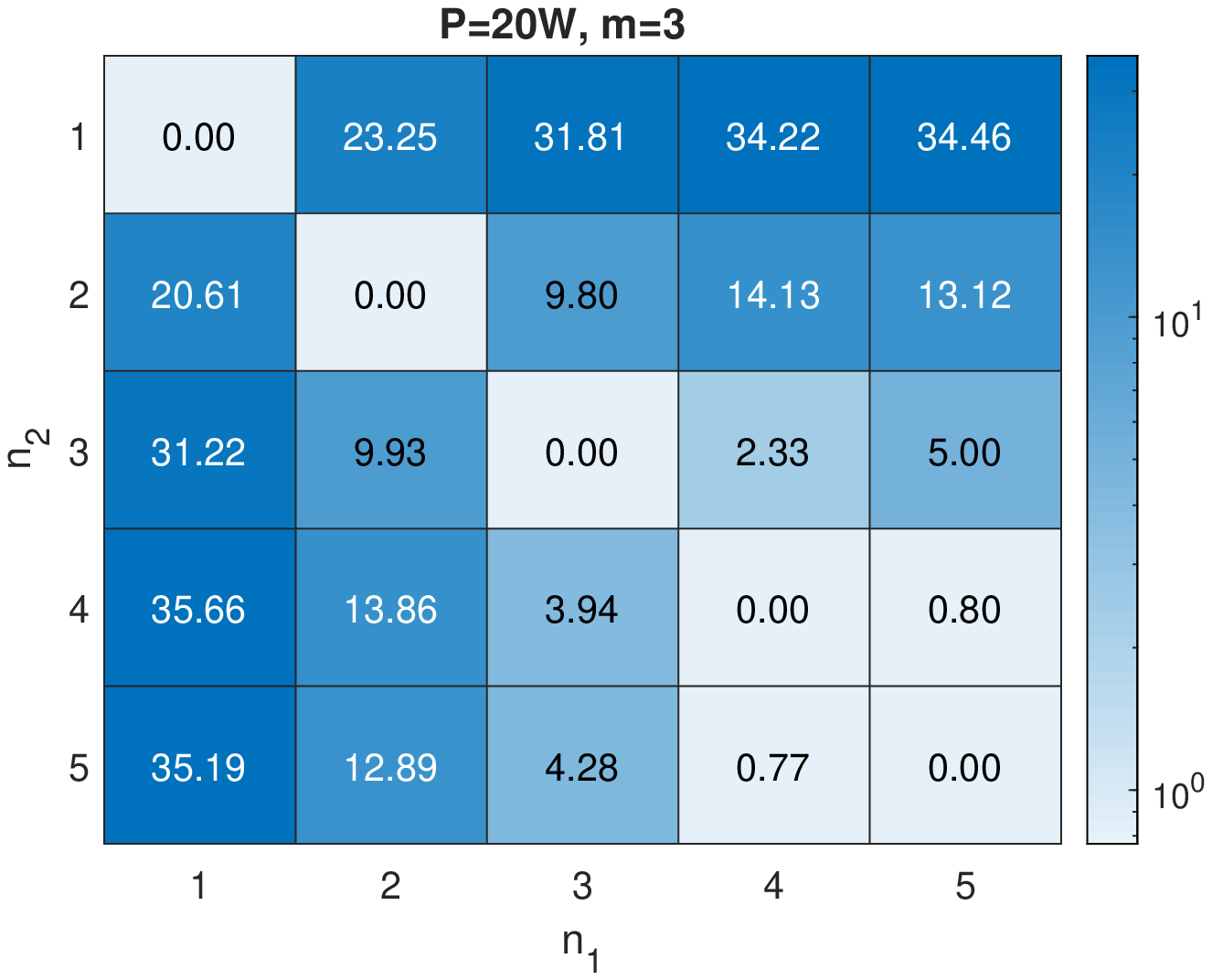}\label{fig_BCmap2}}
	\subfigure
	[The time consumption  of each method.
	]{\includegraphics[width=0.33\textwidth]
		{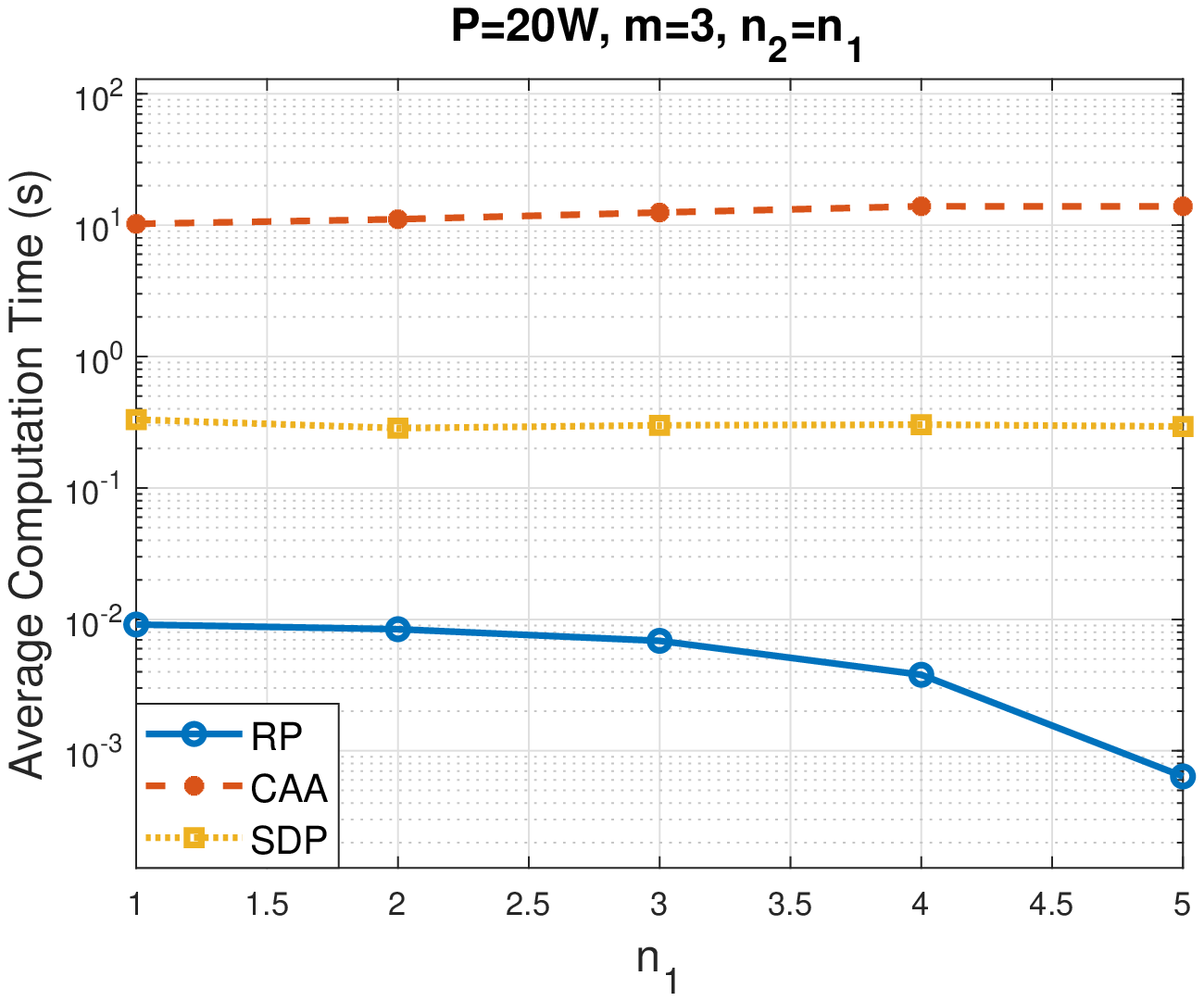}\label{fig_BCcurve1}}
	\caption{Comparisons between the multicasting  transmission rate and 
		computational cost of the RP with CAA and 
		SDP.}\label{fig_BCcomp}
\end{figure*}

\begin{figure}[h]
	\centering
	\includegraphics[width=0.31\textwidth]{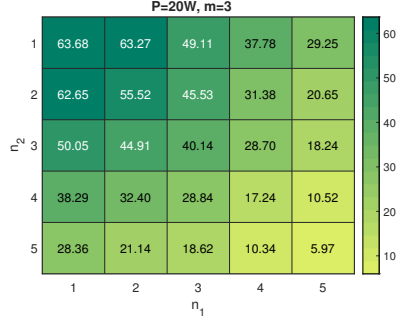}
	\caption{The probability ($\%$) that \textit{Case-3} occurswhen the 
		RP is used for multicasting problem.}
	\label{fig:fig_BC_case3}
\end{figure}

\subsubsection{PHY security 
($\mathcal{O}_4$)}\label{sec_simuA_wt}
{In this subsection, we consider the MIMO wiretap channel. 
The performance of  
RP is  compared with GSVD 
\cite{fakoorian2012optimal} and AO-WF
\cite{li2013transmit}  in 
Fig.~\ref{fig_WTcomp}. 
The 
achievable secrecy rates 
are averaged over 200 random channel realizations, where the channels are 
generated as independent standard Gaussian random variables. 
Fig.~\ref{fig_WTcomp_a} 
and Fig.~\ref{fig_WTcomp_b} represent relative 
improvement defined as 
\begin{subequations}
\begin{align}
&\eta_g=\frac{(R_r-R_g)}{R_g} \times100\%,\label{eq_diff_gsvd}\\
&\eta_a=\frac{(R_r-R_g)}{R_g} \times100\%,\label{eq_diff_ao}
\end{align}
\end{subequations}
in which $\eta_g$ and $\eta_a$ represent the percentages that RP exceeds 
GSVD and AO-WF, i.e., the bluer, the better. $R_r$, $R_g$, and $R_a$ are the 
average secrecy rate 
achieved 
by RP, GSVD, and AO-WF, respectively.  $P=20$W and $m=3$ are set in 
this figure.  Each cell denotes a pair of $n_1$ and $n_2$.
The proposed RP is able  to achieve a better secrecy rate in any 
antenna setting.  There is a noticeable gap between RP 
and GSVD when the eavesdropper has a smaller number of 
antennas. For a larger $n_2$, RP is capable of reaching a higher secrecy rate 
compared to AO-WF. Further illustrations are shown in 
Fig.~\ref{fig_WTcomp_c} 
considering two  cases over 200 channel realizations. The plots show 
the 
average secrecy rates  versus transmit power in the case of $m=3$, 
$n_1=2$, $n_2=1$ and  $m=3$, $n_1=3$, $n_2=5$. We see that RP can 
perform stably and reliably in those cases. Moreover, the average 
time 
costs over all cases in Fig.~\ref{fig_WTcomp_b}  of RP is $27.65$ms which is 
less than  $38.26$ms achieved by 
AO-WF. 


\subsubsection{Multicasting  ($\mathcal{O}_5$)}
For multicasting, we compare RP with the CAA \cite{zhu2012precoder} and 
  standard 
SDP techniques. Here, we 
apply   \texttt{CVX}  
\cite{cvx} 
to realize SDP solutions for multicasting. 

We investigate a variety of combinations of $\{n_1, n_2\}$, and the 
results are 
listed in Fig.~\ref{fig_BCcomp}. Similar to 
\eqref{eq_diff_gsvd}-\eqref{eq_diff_ao}, 
we define the relative improvement factors $\eta_c$ and $\eta_s$ 
representing the 
percentages  that  RP exceeds CAA and SDP,
respectively. 
It can be seen from Fig.~\ref{fig_BCmap1} that  RP 
slightly 
beats  CAA. Besides,  RP outperforms SDP when  $n_1\neq 
n_2$. 
This advantage is especially remarkable when $n_1>n_2=1$ or 
$n_2>n_1=1$ in Fig.~\ref{fig_BCmap2}.
It is worth noting that, in the case of $n_1=n_2$, RP has a 
 very close performance to SDP. This can be found in the diagonal of 
Fig.~\ref{fig_BCmap2}. 

Moreover, the benefits of RP in reducing complexity cannot be 
ignored, which is analyzed in Fig.~\ref{fig_BCcurve1}. Fig.~\ref{fig_BCcurve1} 
compares the time cost when 
$n_1=n_2$ where SDP works well. It can be seen that  RP has the best 
efficiency, while CAA is computationally 
expensive {due to the successive optimization SDP problem.} The 
improvement of time efficiency is partially due to the  
sub-cases we divided in Section~\ref{sec_rot_t7}. In \textit{Case-1} 
and \textit{Case-2}, the solution can be obtained analytically, which is much 
efficient than  \textit{Case-3}. The probability of
\textit{Case-3}  is estimated  by the
Monte Carlo method.  The results are obtained over $20,000$ random 
channels for each specific $n_1$ and $n_2$ with fixed $P=20$W and 
$m=3$. 
As shown in Fig.~\ref{fig:fig_BC_case3}, the probability of \textit{Case-3} is 
reduced with the increase of $n_1$ and $n_2$, which explains the drop 
in the 
time cost of RP in Fig.~\ref{fig_BCcurve1}. 

In summary, RP  provides a unified solution 
for the studied precoding problems. It is feasible for SWIPT. 
Also, it is  
reliable for PHY security and multicasting problems in the variety of 
the number of transmit and receive antennas.


\subsection{Data Set Generation and Training Procedure}
In order to evaluate the performance of the proposed DNN-based precoding,
we generate over two million random realizations of
$\mathbf{H}_1$ and  $\mathbf{H}_2$. Each element of the channels 
follows $\mathcal{N}(0,1)$. From these channel realizations, $2,000,000$ 
channels contribute to the training set and $10,000$ of them are used for 
testing.  
Since the number of 
transmit  antennas $m$ and power $P$ are fixed at the Tx side, 
we set $m=3$ and $P=20$W in all training and test sets. The 
number of 
antennas of UE1 and UE2, i.e., $n_1$ and 
$n_2$,  are randomly chosen from $1$ to $5$ covering most cases of user 
devices. 

For each channel realization in the training and test sets, we generate $13$ 
samples (defined as input-output pairs) corresponding to the 
$K$ objectives 
in Table~\ref{tab_code}. In 
such $K$ samples, each one contains an input feature vector 
according to 
\eqref{eq_inputVec} and a corresponding output vector $\mathbf{q}$ 
formulated as
\eqref{eq_onputVec} given by the solution of the RP 
method. That is, for $\mathcal{O}_1$ and  $\mathcal{O}_2$, 
we have analytical 
solutions given in 
Section~\ref{sec_sys_t1} and \ref{sec_sys_t2}; while for  
$\mathcal{M}_3$ to 
$\mathcal{M}_5$, the   
RP method for each is
given in 
Section~\ref{sec_rot_t3-5} to \ref{sec_rot_t7}. 
Therefore, the training set has $26,000,000$ samples and the test set has 
$130,000$ samples. The details of training and test sets are  summarized in 
Table~\ref{tab_trainingSet}.  
	\begin{table}[htbp]
	\caption{Data Sets Generation.}
	\label{tab_trainingSet}
	\centering
\begin{tabular}{c||ccc||cc}
	\hline
	 Stage       & $m$ & $n_1$ & $n_2$ & 
	\begin{tabular}[c]{@{}c@{}}Number of \\ Channels\end{tabular} & 
	\begin{tabular}[c]{@{}c@{}}Number of \\ Samples\end{tabular} \\ \hline
	Training & 3     & $1,\hdots,5$     & $1,\hdots,5$     & 2,000,000  & 
	26,000,000 \\
	Test          & 3     & $1,\hdots,5$      & $1,\hdots,5$     & 10,000 & 
	130,000      \\ \hline
\end{tabular}
\end{table}

 The training procedure is executed on a single graphical card (NVIDA 
 GeForce GTX 1660Ti) using Adam\cite{kingma2014adam} as  an
 optimization 
 method. 
All training procedures  share the same group of 
hyper-parameters as listed in Table~\ref{tab_param}.  
The learning rate controls how quickly the DNN can change the weights, 
and it  drops by $20\%$ after one epoch in this paper. Mini 
batch size indicates how many samples are considered together for one 
update of the DNN weights. Max epochs denotes the times that all data set 
has 
been taken into the training procedure. 
After 
the
training process, the DNN-based precoding is ready for testing.




\begin{table}[t]
	\caption{Hyper-parameters in Training Procedure.}
	\label{tab_param}
	\centering
	\begin{tabular}{l|c||l|c}
		\hline
		Hyper-parameter        & Value  & Hyper-parameter & Value \\ \hline
		Initial learning rate  & 0.001  & Mini batch size        & 5000  \\
		Learn rate drop factor & 0.8  & Max epochs             & 50     \\
		Learn rate drop period & 1     &  &    \\
		\hline
	\end{tabular}
\end{table}


\subsection{Numerical Results}\label{sec_sub_simu}
 In this part, we evaluate the performance of the  proposed
 unified DNN-based precoding in different antenna settings.  
 The DNN precoder is used in the following way. Two users are equipped with 
 any number of antennas covering from $1$ to $5$, respectively. For a given 
 channel
 pairs $\mathbf{H}_1$ and 
 $\mathbf{H}_2$ with a required mode from Table~\ref{tab_code},  the input 
 layer converts the channels and the mode to sub-feature vectors as 
 \eqref{eq_inputV1}-\eqref{eq_inputV3} and \eqref{eq_inputV0}, 
 respectively.  Then, the output $\mathbf{q}$ in \eqref{eq_onputVec} can be 
 found in the output 
 layer. 
 
 The evaluation of the DNN-precoder contains three 
 metrics: 
\begin{enumerate}
	\item The mean square error (MSE) of  elements in $\mathbf{Q}$ provided 
	by the 	DNN-based precoding;
	\item The performance compared to the conventional methods.
	\item Time 	consumption for each objective.
\end{enumerate}

\subsubsection{MSE of elements of $\mathbf{Q}$} The MSE is evaluated to ensure the feasibility. The MSE is  defined as the 
variance between the ${q}_{i,j}$ given by DNN-based  precoder and the 
one obtained by the corresponding conventional method. 
  The MSE is between   $0.0652$ and $0.0918$  averaged over $130,000$  
  samples in the test   set,   
 which indicates the capability of DNN-based precoding. 
 
 
%

\subsubsection{Performance evaluation}
The performance  of the proposed DNN-based precoder for each objective 
is shown in Fig.~\ref{fig_curve}, where the achievable rate is in 
bit/s/Hz and 
harvested energy in Watt is normalized by the baseband symbol period 
\cite{zhang2013mimo}. Here we  plot the first fifty channel realizations from 
the 
test set for each 
objective. We choose  $\mathcal{O}_3$($60\%$) and 
$\mathcal{O}_3$($20\%$) as 
representatives for SWIPT. In each subfigure, the solid line 
represents the 
achievable rate or harvested energy obtained by conventional methods 
(analytical solutions for $\mathcal{O}_1$ and $\mathcal{O}_2$, 
and RP  for $\mathcal{O}_3$ to $\mathcal{O}_5$),
whereas 
the dashed light-colored line is the result given by the DNN-precoder. It 
can be seen that the results are almost fitting the corresponding 
conventional solutions. 

\begin{figure}[t]
	\centering
	\includegraphics[width=0.65\textwidth]{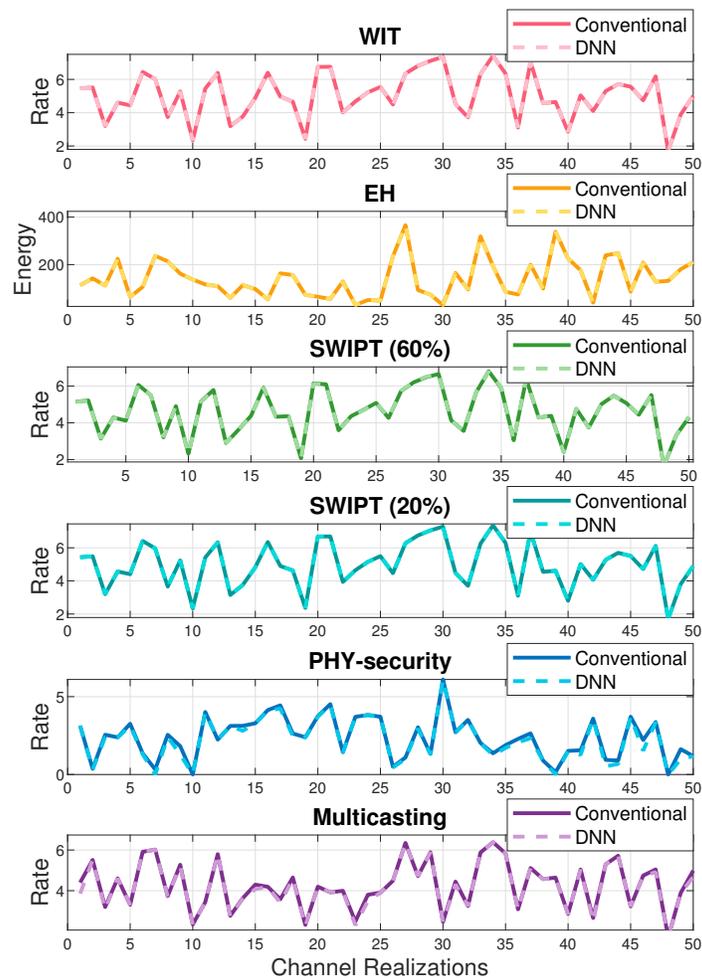}
	\caption{The performance of the proposed DNN-based precoder 
	compared 
		with conventional methods on 6 
		objectives. In this figure, the achievable rate is in bit per second per 
		hertz 
		(bit/s/Hz) and harvested energy is in Watt (W).}
	\label{fig_curve}
\end{figure}

The average achievable rates or harvested energy are reported in 
Table~\ref{tab_avgRate}. The \textit{accuracy}  is defined as the percentage 
of 
DNN-precoder to the conventional methods, i.e., 
\begin{align}\label{eq_achievingRatio}
\zeta_d^{(k)}\triangleq\frac{R_d^{(k)}}{R_c^{(k)}}\times 100 \%,\qquad 
k\in\{1,\hdots,13\},
\end{align}
where $k$ is the  mode index, $R_d^{(k)}$ and $R_c^{(k)}$ are the results 
(achievable rate or harvested energy) of DNN-precoder and the 
 conventional method. Average accuracy is listed in Table~\ref{tab_avgRate}. On average, the accuracy is 
$99.45\%$ among 
all tasks.  The performance of the DNN-precoder could be seen  
the same as the RP method 
except for $\mathcal{O}_4$. 
\begin{table}[t]
	\caption{Average Achievable Rate (bit/s/Hz) or Harvested Energy 
	(W).}
	\label{tab_avgRate}
	\centering
	\begin{tabular}{l||cc||c}
		\hline
		Objective &  Conventional& DNN
		& Accuracy ($\%$)        \\
		\hline
$\mathcal{O}_1$			  &4.9746&4.9733&99.97\\
$\mathcal{O}_2$			  &132.48&132.26&99.84\\
$\mathcal{O}_3$($60\%$)&4.5637&4.5625&99.96\\
$\mathcal{O}_3$($20\%$)&4.9279&4.9261&99.97\\
$\mathcal{O}_4$			  &2.3153&2.1809&94.19\\
$\mathcal{O}_5$			  &4.0142&3.9798&99.14\\
		\hline
	\end{tabular}
\end{table}
It is worth  mentioning that the 
work in
\cite{zhang2019deep}  
is  specifically aimed at $\mathcal{O}_4$ and is only feasible for  
$\{n_1=4, n_2=3\}$ 
and $\{n_1=2, n_2=1\}$. The accuracy in  \cite{zhang2019deep} is 97.71$\%$ 
and 93.72$\%$. In this paper, the DNN-precoder is able to 
achieve 96.94$\%$ 
and 
93.15$\%$ on those two settings of $n_1$ and $n_2$. The slight loss is attributed to much wider antenna settings in this problem as shown in Table~\ref{tab_trainingSet}.


Next, the rate-energy region of SWIPT is demonstrated with more details.
We have arranged nine modes for SWIPT inside the DNN-based precoding, 
which can be
generated as an achievable rate-energy region for any arbitrary channel. In 
Fig.\ref{fig_SW_region}, 
DNN-based solution is compared with conventional methods using the 
channels in \eqref{eq_ch_sw}. Both of 
the methods have been executed for $q=[100\%, 90\%,\hdots,0\%]$. For the
DNN-based precoding, $q=100\%$ is actually  $\mathcal{O}_2$, EH; 
$q=0\%$ is 
obtained by $\mathcal{O}_1$, the WIT.   Those two methods provide 
almost the same 
rate-energy region.

%


\begin{figure}[t]
	\centering
	\includegraphics[width=0.65\textwidth]{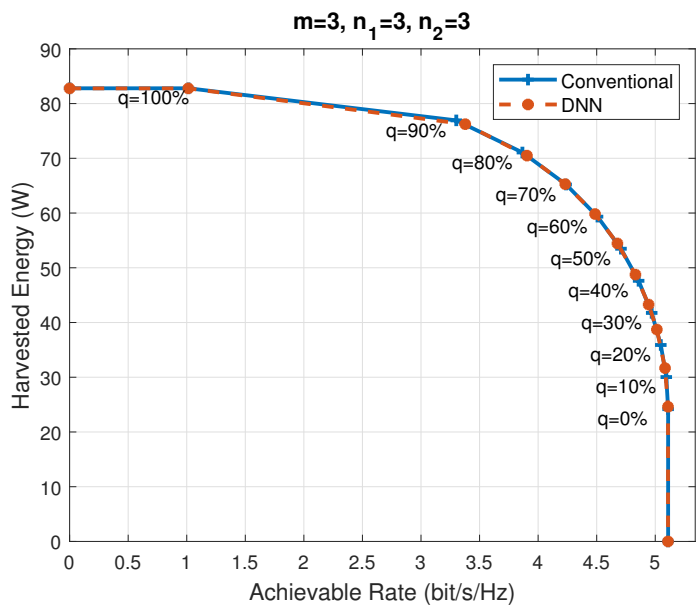}
	\caption{The rate-energy region achievable by DNN-precoder for 
		SWIPT.}
	\label{fig_SW_region}
\end{figure}


\begin{figure}[t]
	\centering
	\includegraphics[width=0.68\textwidth]{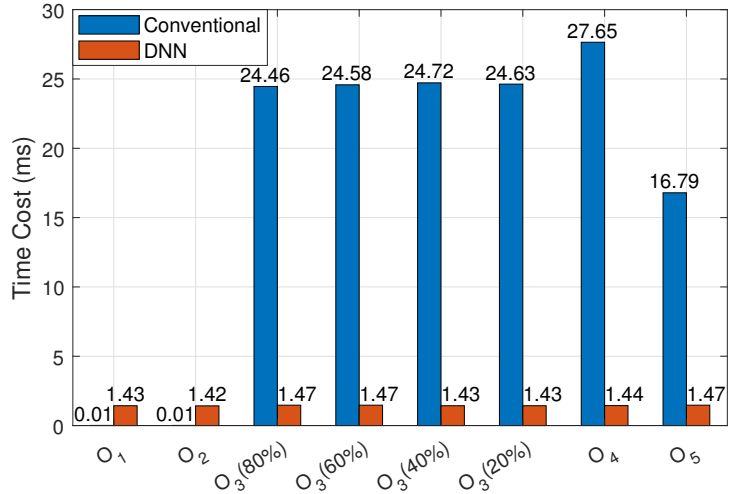}
	\caption{The time cost of each objectives.}
	\label{fig_Time}
\end{figure}

\begin{figure}[t]
	\centering
	\includegraphics[width=0.68\textwidth]{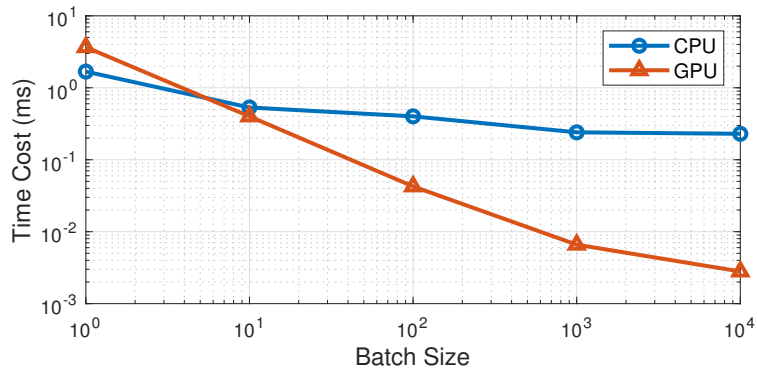}
	\caption{Average execution time over different batch sizes.}
	\label{fig_TimeCGPU}
\end{figure}

\subsubsection{Time consumption}
The average time consumption (averaged over 10,000 channels)  
of the conventional and DNN-based solutions is compared with conventional methods in 
Fig.~\ref{fig_Time}. All objectives are implemented on the 
same CPU, channel by channel. 
For $\mathcal{O}_1$ and 
$\mathcal{O}_2$, conventional 
methods are 
more efficient since they are analytical solutions and are  faster 
than the DNN-based precoder. The advantage of the DNN-based 
precoder 
will appear on other objectives where only numerical solutions exist.
The conventional methods require around 16ms to 27ms to achieve 
solutions 
while  the proposed DNN-based precoding needs less than 1.5ms.  On 
average, it 
saves $91.91\%$ of run-time if we assume that the four objectives occur with 
the same probability.

This indicates that the proposed method is promising for IoT 
applications as 
IoT 
devices which have limited computation capabilities and battery lifetime.
DNN-based precoding is also  good for complicated 
equipment, 
such as base 
stations that serve a large number of users at the 
same time, 
where 
GPUs are affordable for more than one channel.  The average time will 
reduce 
dramatically attributed to the 
parallel computing ability of GPU. We use the \textit{batch size} 
denoting the number of channels or 
objectives processed simultaneously.   Fig.~\ref{fig_TimeCGPU} reveals the 
relation 
between  batch  size and time computation using CPU and GPU. For example, 
when 
batch size is 100, the time cost is 0.4007ms on CPU and 0.0426ms on 
GPU.   
Then, the computational load can be largely reduced.

We understand that generating data sets and training the network  
is time-consuming.
However, these procedures are performed only once and  in an offline 
manner. Once this is done, the DNN-based precoder becomes a matrix multiplication that can be used as long as the assumption on the channels are valid. On the other hand, for the conventional solution the optimization problem corresponding to each objective need to be solved for each input channel independently.  So, the DNN approach shows its advantages in long-term 
usages. 
%
%

\begin{table*}[t]
	\caption{The Accuracy ($\%$) and Time 
		Consumption (ms)  of the 
		DNN-based Precoding 
		with Different Depths and Widths}.
	\label{tab_DEP_WID}
	\centering
	\begin{tabular}{cc||cccccccc||c}
		\hline
		\multirow{2}{*}{Width} &\multirow{2}{*}{Depth} &  
		\multicolumn{8}{c||}{Objective} & {Time Cost}\\  
		& & $\mathcal{O}_1$  &$\mathcal{O}_2$  & 
		$\mathcal{O}_3$($80\%$)& $\mathcal{O}_3$($60\%$) & 
		$\mathcal{O}_3$($40\%$)& $\mathcal{O}_3$($20\%$)& 
		$\mathcal{O}_4$
		& $\mathcal{O}_5$ & (ms)
		\\ 
		\hline
		128 & 10 &99.91&99.52 &99.87&99.94&99.90 
		&99.91&89.65&98.62& 0.94\\
		256 & 6  &99.92&99.58& 99.79& 99.93& 99.94& 99.90& 
		90.69&98.69&1.39\\
		\textbf{256} & \textbf{10} &\textbf{99.97}& \textbf{99.84}& 
		\textbf{99.95}& \textbf{99.97}& \textbf{99.97}& 
		\textbf{99.96}& \textbf{94.19}& \textbf{99.14}&\textbf{1.44}\\
		256 & 14 & 99.98& 99.91& 99.95&  99.98&  99.99& 99.97& 
		96.75&99.15& 2.03\\	
		512 & 10 &99.98& 99.90& 99.97& 99.98& 99.98& 99.98& 
		96.41&99.46&2.39\\
		\hline
	\end{tabular}
\end{table*}

\subsection{The Scale of DNN-based Precoding}
In this part, we evaluate the performance of the DNN-based precoding 
associate with the {depth} (number of hidden layers) and the
 {width} (number of  hidden nodes) of the network. The proposed DNN 
 has ten layers in which nine of them have $256$ hidden nodes and the last 
 one has 
 $128$ hidden nodes. So we define this network as 10 in depth and 256 in 
 width.  
 Other networks 
 are trained and tested in the same way as  we have done in the previous 
 experiments. 
 The performance is 
 listed in Table~\ref{tab_DEP_WID}, including the accuracy in 
 \eqref{eq_achievingRatio} and average time cost. For 	
 $\mathcal{O}_1$ and
 $\mathcal{O}_3$, all of the networks  work well and close to the RP. 
 The performance and the time cost of $\mathcal{O}_2$, 
 	$\mathcal{O}_4$, and 
 	$\mathcal{O}_5$
 are positively related to depth and width.  As a balance of solution 
 quality and time 
 consumption, we choose 
the depth as 10 and width as 256 in this 
paper, even though better performance can be achieved using a deeper and wider network.

\section{Conclusion}\label{sec_conclu}
In this paper, a unified DNN-based precoder has been proposed for green, secure wireless transmission in two-user MIMO systems. Specifically, WIT, EH, 
SWIPT, PHY security, and multicasting problems have been considered. 
We first use rotation-based precoding to derive the transmit 
covariance matrix for the above problems from which the DNN learns. The overall performance of the rotation-based precoder is  better than the existing methods for SWIPT, PHY security, and multicasting.
These conventional methods based on the mathematical models are  
used for data set generation and training procedure of DNN. 
Next, a DNN-based precoder is designed to unify the 
solutions for different objectives.  This  DNN-based precoding 
can effectively optimize 
all 
objectives at the same time. In terms of achievable rates and harvested energy, The performance of the unified 
DNN-based precoder is 
similar to the method it learns from, whereas its  time cost is 
substantially 
lower than the conventional iterative solutions. Due to its lower 
computational complexity and its 
high flexibility, the proposed precoding is suitable for emerging existing 
applications, where  the low-latency  and low-complexity devices are necessary.

\balance
\typeout{}
\bibliography{REF_commu_v1.0}
\bibliographystyle{ieeetr}

\end{document}